\documentclass[letterpaper, 10 pt, journal, twoside]{IEEEtran}
\ifCLASSINFOpdf
\else
\fi
\def\BibTeX{{\rm B\kern-.05em{\sc i\kern-.025em b}\kern-.08em
    T\kern-.1667em\lower.7ex\hbox{E}\kern-.125emX}}
\usepackage{balance}
\usepackage{amssymb}
\usepackage{amsmath,amsfonts}
\usepackage{algorithmic}
\usepackage{algorithm}
\usepackage{array}
\usepackage{textcomp}
\usepackage{stfloats}
\usepackage{url}
\usepackage{graphicx}
\usepackage{cite}
\newlength{\FigWidth}
\usepackage[caption=false]{subfig}
\usepackage[font=small,skip=2pt]{caption}
\usepackage{xcolor}
\usepackage{bm}
\usepackage{here}
\usepackage{amsthm}
\newtheorem{theorem}{Theorem}[section]

\newtheorem{remark}{Remark}

\newtheorem{assumption}{Assumption}

\usepackage{graphicx}
\usepackage{textcomp}
\usepackage{comment}
\begin{document}
\title{Certified Coil Geometry Learning for Short-Range Magnetic Actuation and Spacecraft Docking Application}
\author{Yuta Takahashi$^{1}$, Hayate Tajima$^{2}$,
Shin-ichiro Sakai$^{3}$
\thanks{Manuscript received: November, 17, 2025; Revised February, 14, 2026; Accepted March, 27, 2026.}
\thanks{This paper was recommended for publication by Editor Cosimo Della Santina upon evaluation of the Associate Editor and Reviewers’ comments. This work was partially supported by the JAXA, Space Strategy Fund (JPJXSSF24MS09003). Code: https://github.com/stateofyuta-y-cl/CGL.}%
\thanks{$^{1}$Yuta Takahashi is with the Department of Mechanical Engineering, Institute of Science Tokyo, Tokyo, Japan. He is with Satellite Research and Development, Interstellar Technologies Inc., Hokkaido, Japan {\tt\footnotesize stateofyuta@gmail.com}}%
\thanks{$^{2}$Hayate Tajima is with the Department of Advanced Energy, The University of Tokyo, Chiba, Japan {\tt\footnotesize liverpool.bayern.0821@gmail.com}}
\thanks{$^{3}$Shin-ichiro Sakai is with the Department of Spacecraft Engineering, Japan Aerospace Exploration Agency, Kanagawa, Japan {\tt\footnotesize sakai@isas.jaxa.jp}}
\thanks{Digital Object Identifier (DOI): see top of this page.}
}
\markboth{IEEE Robotics and Automation Letters. Preprint Version. Accepted March, 2026 (DOI: 10.1109/LRA.2026.3685510)}
{Takahashi \MakeLowercase{\textit{et al.}}: Certified Coil Geometry Learning for Short-Range Magnetic Actuation and Spacecraft Docking Application} 
\maketitle
\begin{abstract}
This paper presents a learning-based framework for approximating an exact magnetic-field interaction model, supported by both numerical and experimental validation. High-fidelity magnetic-field interaction modeling is essential for achieving exceptional accuracy and responsiveness across a wide range of fields, including transportation, energy systems, medicine, biomedical robotics, and aerospace robotics. In aerospace engineering, magnetic actuation has been investigated as a fuel-free solution for multi-satellite attitude and formation control. Although the exact magnetic field can be computed from the Biot–Savart law, the associated computational cost is prohibitive, and prior studies have therefore relied on dipole approximations to improve efficiency. However, these approximations lose accuracy during proximity operations, leading to unstable behavior and even collisions. To address this limitation, we develop a learning-based approximation framework that faithfully reproduces the exact field while dramatically reducing computational cost. \textcolor{black}{This framework directly derives a coefficient matrix that maps inter-satellite current vectors to the resulting forces and torques, enabling efficient computation of control current commands.} The proposed method additionally provides a certified error bound, derived from the number of training samples, ensuring reliable prediction accuracy. The learned model can also accommodate interactions between coils of different sizes through appropriate geometric transformations, without retraining. To verify the effectiveness of the proposed framework under challenging conditions, a spacecraft docking scenario is examined through both numerical simulations and experimental validation.
\end{abstract}
\begin{IEEEkeywords}
Machine Learning for Robot Control, Space Robotics and Automation, Model Learning for Control.
\end{IEEEkeywords}
\IEEEpeerreviewmaketitle
\section{Introduction}
\IEEEPARstart{E}{xact} magnetic-field interaction modeling plays a central role across diverse domains that require high accuracy and responsiveness. In transportation and infrastructure, it supports linear motors and maglev systems by coordinating propulsion, levitation, and guidance. 
The energy sector likewise depends on precise inductance estimation for wireless power transfer, drawing on analytical formulations \cite{khan2018accurate}. 
In medicine, triaxial coil systems enable multichannel transcranial magnetic stimulation, enhancing the spatial selectivity of neural excitation \cite{de20213a3axis}. Magnetic fields have been used in biomedical robotics to steer microrobots through internal organs, facilitating targeted drug delivery and minimally invasive procedures \cite{ze2022soft}. In the aerospace sector, Earth's magnetic environment has been exploited to control spacecraft attitude using magnetorquers (MTQs). Conventional actuators, such as thrusters, generate a plume that causes sensor contamination and disturbances in proximity operations. In contrast, magnetic-field interaction control mitigates these issues, enhancing mission flexibility and longevity. Several studies apply this actuation to proximity operations, including non-contact micro-vibration suppression \cite{shibata2022contactless}, 
debris removal \cite{fabacher2017guidance}, formation keeping \cite{takahashi2025distance}, and satellite docking \cite{foust2018ultra,tajima2023study}. These highlight magnetic-field modeling as a unifying technology for control, sensing, and actuation.

\begin{figure}[!t]
\centering
{\includegraphics[width=0.975\FigWidth]{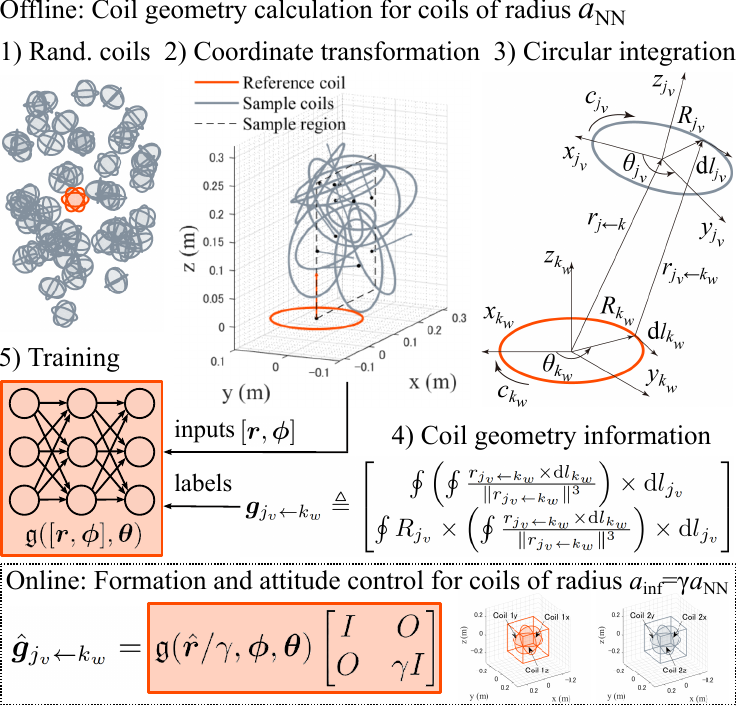}}
\caption{Coil geometry learning to predict the magnetic interaction: \textcolor{black}{Random coils are generated around a reference coil and associated Biot–Savart–based coefficients are learned, enabling efficient online prediction and reuse across different coil diameters without retraining.}}
\label{CGL_fig:coil_geometry_learning}
\end{figure}
The aerospace community has been active in developing a theoretical framework for controlling magnetic-field interactions to achieve multiple objectives. Previous studies have considered magnetic-field interactions in space to control the relative positions of various satellites with extremely high precision. They are referred to as electromagnetic formation flight \cite{porter2014demonstration,takahashi2025experimental,schweighart2006electromagnetic,takahashi2022kinematics,tajima2023study,takahashi2025noda_mmh,abbasi2022decentralized} with demonstrations on the International Space Station \cite{porter2014demonstration}. Since this system is naturally underactuated for swarm control \cite{takahashi2022kinematics}, previous studies extend controllability through time-integrated current control, such as the alternating current (AC) method 
\cite{abbasi2022decentralized,takahashi2022kinematics}. 
Recent studies investigate the six-degree-of-freedom control of a large number of satellites through AC control \cite{takahashi2022kinematics} and demonstration \cite{takahashi2025noda_mmh} with a learning-aided technique \cite{takahashi2024neural}. These lead to a new class of commercial and scientific missions, such as the fuel-free on-orbit construction of space structures \cite{shim2025feasibility}. 

However, the conventional magnetic interaction model is invalid for short-range interactions, and predicting the exact interaction is impractical in real-time due to the high computational burden. The exact magnetic interaction model developed for formation control \cite{schweighart2006electromagnetic} includes the double circular integration using coil geometry and the relative coil states. This is calculated based on the Biot–Savart law, and the associated computational cost is prohibitive. To avoid its cost, existing research primarily uses the far-field approximation \cite{schweighart2006electromagnetic}, assuming that the distance between coils is significantly greater than the coil radius. While this model reduces computational cost, it becomes less accurate as the distance between the coils decreases. Consequently, the desired control cannot be achieved at short distances without proper strategy, leading to inaccuracies, instability, and, in the worst case, satellite collision \cite{tajima2023study}. Therefore, real-time calculations were always a trade-off between accuracy and computational time. 

To this end, this study proposes a learning-based framework to predict the exact magnetic field interaction with the certificated model error bound. The effectiveness of our approaches is demonstrated through numerical and experimental methods in the context of docking. For these validations, we apply our learning framework to a decentralized current allocation without inter-agent communication, extending the far-field model strategy \cite{takahashi2024neural}. The overall structure of our study is as follows: Section~\ref{CGL_Preliminaries} formulates a multilayer perceptron model and the exact magnetic model of a one-axis coil to prepare for the derivation of our framework. Section~\ref{CGL_Problem_Formulation_and_Decentralized_Docking} derives an interaction model using a three-axis coil referred to as ``coil geometry information'' and a communication-free decentralized current calculation for the exact magnetic-field model. Finally, our learning-based coil-specific geometry modeling framework is presented in Section~\ref{CGL_Learning_based_Coil_Specific_Geometry_Approximation}, along with generalization methods for coils of arbitrary radii and error-bound certification. Section~\ref{CGL_Numerical_Validation} conducts the offline training and validates the proposed learning-based current calculation for satellite docking through numerical simulations and experimental comparison with the conventional approximation model. 
\section{Preliminaries}
\label{CGL_Preliminaries}
This section introduces the magnetic field interaction model for multi-agent control as preliminaries. Our mathematical notation for vectors and coordinates is as follows: $\{\boldsymbol{a}\}$ is defined as basis vector of an arbitrary frame $(A)$ such that $\{\boldsymbol{a}\}=\left\{\boldsymbol{a}_x^{\mathrm{T}}, \boldsymbol{a}_y^{\mathrm{T}}, \boldsymbol{a}_z^{\mathrm{T}}\right\}^{\mathrm{T}}$ and we define $p^a$ component of an arbitrary vector $\bm{p}\in\mathbb{R}^3$ in this frame, i.e., $\bm{p}=\{\boldsymbol{a}\}^\top p^a
=\{\boldsymbol{b}\}^\top p^b=\{\boldsymbol{b}\}^\top C^{B/A}p^a$ where $C^{B/A}\in\mathbb{R}^{3\times 3}$ is the coordinate transformation matrix from frame (A) to (B). We define the Lipschitz constant $\|f\|_{{\mathrm{Lip}}}$ for a general differentiable function $f$ as $\forall {x}_1,{x}_2:
{\left\|f_{({x}_1)}-f_{\left({x}_2\right)}\right\|_2}/{\left\|{x}_1-{x}_2\right\|_2} \leq\|f\|_{\mathrm{Lip}}= \sup _{{x}} \sigma(\nabla f_{({x})})$ and $\left\|f_1 \circ f_2\right\|_{\text {Lip }} \leq\left\|f_1\right\|_{\text {Lip }}\left\|f_2\right\|_{\text {Lip }}$.
\subsection{Multilayer Perceptron Model and Error Bound}
\label{CGL_certified_error_bound}
This subsection introduces the conservative error bound between true value and approximation values generated by the Multilayer Perceptron (MLP) model. The MLP represents the functional mapping from inputs $x$ into outputs $\mathfrak{g}_{(\bm{X}^{(j)}, \bm{\theta})}$ such as a ($L$+1)-layer neural network:
\begin{equation*}
\mathfrak{g}_{(\bm{X}^{(j)}, \bm{\theta})}= W^{(L+1)}\phi(\cdots \phi(W^{(1)} X^{(j)}+b^{(1)})\cdots )+b^{(L+1)}
\end{equation*}
where the activation function $\phi(\cdot)$ and the MLP paramters $\theta$ include the weights $\theta_w = W^{(1)},\ldots, W^{(L+1)}$, and the bias $\theta_b = b^{(1)},\ldots, b^{(L+1)}$. Since the spectral norm of each layer is $\|W^{(l)}\mathbf{x}+b^{(l)}\|_{\text {Lip }}=\sup _{\mathbf{x}} \sigma(\nabla (W^{(l)}\mathbf{x}+b^{(l)}))=\sigma(W^{(l)})$, $\|\mathfrak{g}\|_{\text {Lip }}$ is bounded as $\|\mathfrak{g}\|_{\text {Lip }} \leq
\|\phi\|_{\text {Lip }}^L\prod_{l=1}^{L+1} \sigma\left(W^{(l)}\right)$ \cite{miyato2018spectral}. Then, we obtain the Liptchitz constant of the error function $\mathsf{e}_{g/\mathfrak{g}(x)}=\|g_{x}-\mathfrak{g}_{x}\|$ as $(L_g + L_{\mathfrak{g}})$ since we have
$$
\|\mathsf{e}_{g/\mathfrak{g}(x)}-\mathsf{e}_{g/\mathfrak{g}(y)}\|=\|(g_{x}-g_y)-(\mathfrak{g}_{x}-\mathfrak{g}_{y})\|\leq (L_g + L_{\mathfrak{g}})\|x-y\|
$$
where the Liptchitz constants $L_g$ for the true function and $L_{\mathfrak{g}}$ for the MLP model. The upper bound of learning error, i.e., $\sup_{x\in \mathcal{X}}\|g_{x}-\mathfrak{g}_{x}\|$, for a compact set $\mathcal{S}_{R}$ is bounded as
\begin{equation}
\label{CGL_upper_bound_learning_error}
\sup_{x\in \mathcal{S}_{R}}\|g_{x}-\mathfrak{g}_{x}\|\leq\sup_{x'\in\mathcal{T}_{R}}\| g_{x'}-\mathfrak{g}_{x'}\|+(L_g + L_{\mathfrak{g}})\rho_{\mathcal{S}_{R}}
\end{equation}
where the covering radius $\rho_{\mathcal{S}_{R}}=\sup_{x\in\mathcal{S}_{R}}\min_{x'\in \mathcal{T}_{R}}\|x-x'\|$.
\subsection{Electromagnetic Interaction Model of One-Axis Coil}
\label{CGL_Electromagnetic_Interaction_Model_of_One_Axis_Coil}
\begin{figure}[tb!]
  \centering
  \begin{minipage}{0.33\FigWidth}
    \centering
    {%
      \includegraphics[width=\linewidth]{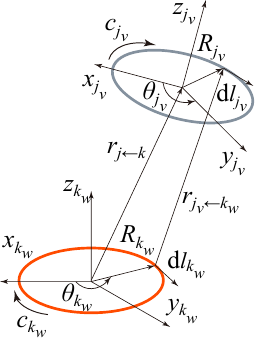}%
    }
  \end{minipage}
  \begin{minipage}{0.32\FigWidth}
    \centering
      \begin{minipage}[b]{\columnwidth}
  \centering
\includegraphics[width=0.95\linewidth]{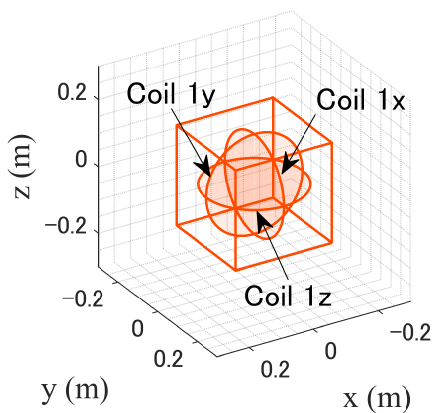}
  \end{minipage}
\begin{minipage}[b]{\columnwidth}
  \centering
\includegraphics[width=0.95\linewidth]{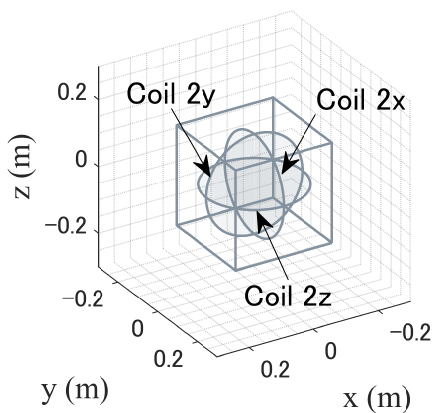}
  \end{minipage}
  \end{minipage}
  \begin{minipage}{0.32\FigWidth}
    \centering
      \begin{minipage}[b]{\columnwidth}
  \centering
\includegraphics[width=0.95\linewidth]{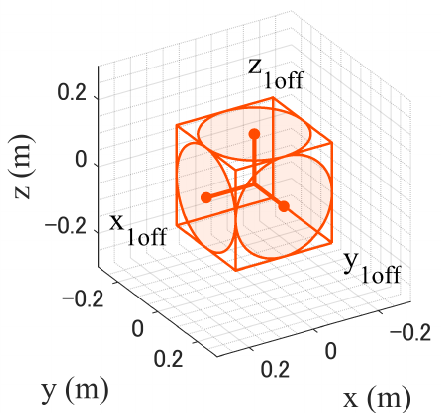}
  \end{minipage}
\begin{minipage}[b]{\columnwidth}
  \centering
\includegraphics[width=0.95\linewidth]{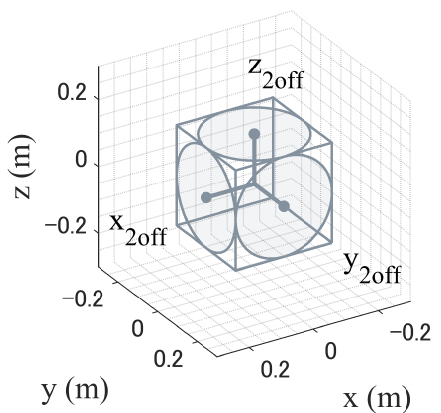}
  \end{minipage}
  \end{minipage}
  \caption{%
    The definition of circulant integration in (\ref{CGL_circulant_integration_term}) and the coil model of target and chaser with offset in subsection~\ref{CGL_Three_Axis_Coil_Geometry_Information_with_Coil_Offset}. \textcolor{black}{The right-column coils are offset from the body center by $r_{j_{v}\mathrm{off}}$ in (\ref{CGL_near_force_and_torque}).}
  }
  \label{CGL_circulant_integration}
\end{figure}

This subsection derives the magnetically interacting model between one-axis coils. We model the $k$th one-axis coil as a circular air-core coil and define its dipole moment as 
$\mu_{k}=N_{t}A\mathsf{c}_k \bm n$
where the number of coil turns $N_t$, the area enclosed by the coil $A=\pi a_k^2$ with the coil radius $a_k$, the current strength $\mathsf{c}_{k}(t)$, and $\bm n$ is the unit vector perpendicular to the plane of the coil. 
The exact magnetic field model $B(r_{j\leftarrow k})$ \cite{schweighart2006electromagnetic} using the well-known Biot-Savart law is
$$
B(r_{j\leftarrow k})=\frac{\mu_0 c_k}{4 \pi} \oint \frac{{\mathrm{d}l_k} \times r_{j_{v}\leftarrow k_w(\theta_{j_v},\theta_{k_w})}}{\|r_{j_{v}\leftarrow k_w(\theta_{j_v},\theta_{k_w})}\|^3}\in\mathbb{R}^3
$$ 
where the magnetic permeability $\mu_0$, the coil element $\mathrm{d}l_k\in\mathbb{R}^3$, and the distance between coil elements $r_{j_{v}\leftarrow k_w(\theta_{j_v},\theta_{k_w})}=r_{j\leftarrow k}+C^{k/j}R_j-R_k$, the coil radius vector $R_{k(\varphi)}=a_k[\cos\varphi,\sin\varphi,0]^\top$, the $k$th coil element vector $\mathrm{d}\ell_{k(\varphi)}=(\mathrm{d}R_{k(\varphi)}/\mathrm{d}\varphi)\mathrm{d}\varphi$. The electromagnetic force and torque experienced by the $j$th one-axis coil due to the magnetic field generated by the $k$th one-axis coil $u^{\mathrm{1axis}}_{j\leftarrow k}=[f_{j\leftarrow k};
\tau_{j\leftarrow k}]$ are
\begin{equation*}
u^{\mathrm{1axis}}_{j\leftarrow k}
=\frac{\mu_0\mu_{k}\mu_{j}}{4 \pi A^2}\ 
\begin{bmatrix}
\int_{\theta_j=0}^{\theta_j=2 \pi}
\int_{\theta_k=0}^{\theta_k=2 \pi} \frac{r_{j_{v}\leftarrow k_w} \times {\mathrm{d}l}_{k}}{\|r_{j_{v}\leftarrow k_w}\|^3} \times {\mathrm{d}l}_{j}\\
\int_{\theta_j=0}^{\theta_j=2 \pi}
R_{j} \times \int_{\theta_k=0}^{\theta_k=2 \pi} \frac{{{r}_{j_v\leftarrow k_w}} \times {\mathrm{d}l}_{k}}{\|r_{j_v\leftarrow k_w}\|^3} \times {\mathrm{d}l}_{j}
\end{bmatrix}
\end{equation*}
where Fig.~\ref{CGL_circulant_integration} provides an overview of this integration. 
\subsection{``Far-Field'' Approximation Model}
\label{CGL_appendix_far_field_approximation}
``Far-field'' model \cite{schweighart2006electromagnetic} provides a computationally friendly approximate magnetic field. 
We assume the size of the coil loop is much smaller than the relative distance, i.e., $|R_{i}|\ll|r_{j\leftarrow k}|$. Then, we get ${1}/{|r|}\approx 1/{|s_{ij}|}+{s_{ij} R_{i}}/{|s_{ij}|^3}$. This approximation derives a dipole model of the multilayer coil as $j$th magnetic moment $\mu_{j}(t)\in\mathbb{R}^3$ [Am$^2$] at time $t$
\begin{equation}
\label{CGL_dipole_moment}
\bm{\mu}_{j(t)}=\{\bm{b}\}^\top \mu^b_{j(t)},\quad \mu^b_{j(t)}\triangleq 
N_{t}A
\begin{bmatrix}
    \mathsf{c}_{jx(t)};
    \mathsf{c}_{jy(t)};
    \mathsf{c}_{jz(t)}
\end{bmatrix}
\end{equation}
where the current strength $\mathsf{c}_{j(x,y,z)}(t)$ along with $x,y,z$-axis in $j$th body-fixed frame. Its magnetic field $B_k({\mu}_k, {r}_{j k})=\frac{\mu_0}{4 \pi d_{jk}^3}(3M_k\mathsf{e}_r -{\mu}_k)$ where $M_k=\bm{\mu}_k  \mathsf{e}_r$, $d_{jk}=\left\|\bm{r}_{j k}\right\|$, and $\mathsf{e}_r={\bm{r}_{j k}}/{d_{jk}}$. 
This also simplifies the magnetic field interaction model $\bm{f}_{j\leftarrow k}=\nabla({\mu}_j  {B}_k)$ and $\bm{\tau}_{j\leftarrow k}=\bm{\mu}_j \times B_k$ where
$\bm{f}_{j\leftarrow k}=\frac{3 \mu_0}{4 \pi d^4}\left(({\bm{\mu}_k  \bm{\mu}_j}-5 M_k M_j)\mathsf{e}_r+M_k \bm{\mu}_j+M_j\bm{\mu}_k\right)$.
\section{Coil Geometry Information and Decentralized Dipole Allocation}
\label{CGL_Problem_Formulation_and_Decentralized_Docking}
\subsection{Three-Axis Coil Geometry Information with Coil Offset}
\label{CGL_Three_Axis_Coil_Geometry_Information}
We extend the interaction model in subsection~\ref{CGL_Electromagnetic_Interaction_Model_of_One_Axis_Coil} to a three-axis coil with offset illustrated in Fig.~\ref{CGL_circulant_integration} to define the ``coil geometry information vector.'' 
We assume \textcolor{black}{the $j$th and $k$th} agents have identical circular triaxial air-core coils whose radius is $a=a_j=a_k$, and the $j$th dipole moment $\mu_{j(t)}\in\mathbb{R}^3$ at time $t$ in (\ref{CGL_dipole_moment}). 
We define the coil geometry vector of 3-axis agents between each axes $v,w=(x,y,z)$ as ${\bm{g}}_{j_{v}\leftarrow k_w}\in\mathbb{R}^6$ that includes the circulant integration term:
\label{CGL_Three_Axis_Coil_Geometry_Information_with_Coil_Offset}
\begin{equation}
\label{CGL_circulant_integration_term}
\begin{aligned}
&{\bm{g}}_{j_{v}\leftarrow k_w}\ (r_{j\leftarrow k},C_j,a)\triangleq
\begin{bmatrix}
{\bm{g}}^{\mathrm{pos}}_{j_{v}\leftarrow k_w}{(r_{j\leftarrow k},C_j,a)}\\
{\bm{g}}^{\mathrm{rot}}_{j_{v}\leftarrow k_w}{(r_{j\leftarrow k},C_j,a)}
\end{bmatrix}\\
&\triangleq
\begin{bmatrix}
\int_{\theta_{j_{v}}=0}^{\theta_{j_{v}}=2 \pi}\left({\bm{g}}_{j_{v}\leftarrow k_w(\theta_{j_v})}^{\mathrm{inner}}\right)\times {\mathrm{d}l}_{j_{v}(\theta_{j_v})}\\
\int_{\theta_{j_{v}}=0}^{\theta_{j_{v}}=2 \pi} R_{j_{v}(\theta_{j_v})} \times \left(\left({\bm{g}}_{j_{v}\leftarrow k_w(\theta_{j_v})}^{\mathrm{inner}}\right) \times {\mathrm{d}l}_{j_{v}(\theta_{j_v})}\right)\\
\end{bmatrix}
\end{aligned}
\end{equation}
where the inner circular integration term ${\bm{g}}_{j_{v}\leftarrow k_w (\theta_{j_v})}^{\mathrm{inner}}\in\mathbb{R}^3$ is
$$
{\bm{g}}_{j_{v}\leftarrow k_w (\theta_{j_v})}^{\mathrm{inner}}\triangleq\int_{\theta_{k_w}=0}^{\theta_{k_w}=2 \pi} \frac{{{r}_{j_{v}\leftarrow k_w(\theta_{j_v},\theta_{k_w})}} \times {\mathrm{d}l}_{k_{w}(\theta_{k_w})}}{\|r_{j_{v}\leftarrow k_w(\theta_{j_v},\theta_{k_w})}\|^3}
$$
This defines the matrix $G_{j\leftarrow k}\in\mathbb{R}^{6\times 9}$

\begin{equation}
\label{CGL_near_force_and_torque}
\left\{
\begin{aligned}
{\bm{g}}_{j_{v}\leftarrow k_w\mathrm{off}}&=
\begin{bmatrix}
    I_3&O_3\\
\left[r_{j_{v}\mathrm{off}}\right]_\times&I_3
\end{bmatrix}{\bm{g}}_{j_{v}\leftarrow k_w}\\
{\bm{g}}_{j\leftarrow k_w}&=
\begin{bmatrix}
{\bm{g}}_{j_{x}\leftarrow k_w\mathrm{off}}&
{\bm{g}}_{j_{y}\leftarrow k_w\mathrm{off}}&
{\bm{g}}_{j_{z}\leftarrow k_w\mathrm{off}}
\end{bmatrix}\\
G_{j\leftarrow k}&=A^{-2}
\begin{bmatrix}
{\bm{g}}_{j\leftarrow k_x}&{\bm{g}}_{j\leftarrow k_y}&{\bm{g}}_{j\leftarrow k_z}
\end{bmatrix}
\end{aligned}
\right.
\end{equation}
Finally, the $j$th input due to the magnetic field generated by the $k$th agent $u_{j\leftarrow k}=[f_{j\leftarrow k};\tau_{j\leftarrow k}]$ and the total input on the $j$-agents $u_{j}=[f_{j};\tau_{j}]$ due to the neighbor agents $\mathcal{N}_j$ are
\begin{equation}
\label{CGL_near_field_electromagnetic_interaction_model}
u_{j\leftarrow k}=\frac{\mu_0}{4\pi}G_{j\leftarrow k} \left(\mu^b_k\otimes \mu^b_j\right),\quad u_{j}=\sum_{k\in\mathcal{N}_j}
u_{j\leftarrow k}.
\end{equation}
\subsection{Decentralized Dipole Allocation Using Alternating Current}
We derive the control currents in a decentralized manner for simultaneous control of relative position and absolute attitude by extending the far-field study \cite{takahashi2024neural}. We first summarize alternating current (AC) modulation for multi-agent control.
\begin{assumption}
The $j$th agent applies a periodic $\mu_j(t)$ for coils:
$$
\begin{aligned}
\mu_j(t)={s}_{j}\sin \omega t+{c}_{j}\cos \omega t.
\end{aligned}
$$
where $\omega$ [rad/s] is the angular frequency and sufficiently higher than the system dynamics time constant, i.e., $\frac{\mathrm{d} \overline{x}}{\mathrm{d} t}=\int_0^T \frac{\mathrm{d} x}{\mathrm{d} \tau}(t, \tau) \frac{\mathrm{d} \tau}{T} +\epsilon$ and $\epsilon\ll 1$.
\end{assumption}
\noindent
We calculate ${s}_{j}$ and ${c}_{j}$ so that the average value of the electromagnetic force ${f}^{\mathrm{avg}}_{j\leftarrow k}$ and torque $\tau^{\mathrm{avg}}_{j\leftarrow k}$ matches the control command values ${f}_{jc},\tau_{jc}$ \cite{ takahashi2022kinematics,abbasi2022decentralized}, i.e.,
$$
\sum_{k\in\mathcal{N}_j}
\begin{bmatrix}
\int_T{f}_{j\leftarrow k}(u)\frac{\mathrm{d}u}{T}\\
\int_T{\tau}_{j\leftarrow k}(u)\frac{\mathrm{d}u}{T}
\end{bmatrix}
\triangleq
\sum_{k\in\mathcal{N}_j}
\begin{bmatrix}
{f}^{\mathrm{avg}}_{j\leftarrow k}\\
{\tau}^{\mathrm{avg}}_{j\leftarrow k}
\end{bmatrix}
\equiv
\begin{bmatrix}
{f}_{jc}\\
{\tau}_{jc}
\end{bmatrix}.
$$
Based on (\ref{CGL_near_field_electromagnetic_interaction_model}), the averaged electromagnetic force and torque between two agents equipped with three-axis coils are
\begin{equation}
\label{CGL_near_field_interaction_model}
\begin{aligned}
\begin{bmatrix}
{f}^{\mathrm{avg}}_{j\leftarrow k}\\
{\tau}^{\mathrm{avg}}_{j\leftarrow k}
\end{bmatrix}
&\triangleq\int_T\frac{\mu_0}{4\pi}G_{j\leftarrow k}
(\mu^{b}_{k}(u)\otimes \mu^{b}_{j}(u))\frac{\mathrm{d}u}{T}\\
&\approx \frac{1}{2}\frac{\mu_0}{4\pi}G_{j\leftarrow k}\left(s^b_{k}\otimes s^b_{j}+c^b_{k}\otimes c^b_{j}\right)
\end{aligned}
\end{equation}
Then, the decentralized derivation of control currents that extends the far-field model one \cite{takahashi2024neural} is
\begin{equation}
\label{CGL_decentralized_current_allocation}
\left\{
\begin{aligned}
 &
\begin{bmatrix}
    s_{j}^a\\
    c_{j}^a
\end{bmatrix}=
\left(\frac{\mu_0}{8\pi}G_{j\leftarrow k}^a \left([s_{k}^a,c_{k}^a]\otimes E_3\right)\right)^{-1}
\begin{bmatrix}
{f}^{a(\mathrm{avg})}_{j\leftarrow k}\\
{\tau}^{a(\mathrm{avg})}_{j\leftarrow k}
\end{bmatrix}\\
 &
\begin{bmatrix}
    s_{k}^a\\
    c_{k}^a
\end{bmatrix}=
\left(\frac{\mu_0}{8\pi}G_{j\leftarrow k}^a \left(E_3\otimes[s_{j}^a,c_{j}^a]\right)\right)^{-1}
\begin{bmatrix}
{f}^{a(\mathrm{avg})}_{j\leftarrow k}\\
{\tau}^{a(\mathrm{avg})}_{j\leftarrow k}
\end{bmatrix}
\end{aligned}
\right.
\end{equation}
where we use $\left(\mathbf{a}\otimes \mathbf{b}+\mathbf{c}\otimes \mathbf{d}\right)=\left([\mathbf{a},\mathbf{c}]\otimes E_3\right)[\mathbf{b};\mathbf{d}]=\left(E_3\otimes [\mathbf{b},\mathbf{d}] \right)[\mathbf{a};\mathbf{c}]$ for arbitrary vectors $\mathbf{a},\mathbf{b},\mathbf{c},\mathbf{d}$.
\section{Learning-based Coil Geometry Approximation}
\label{CGL_Learning_based_Coil_Specific_Geometry_Approximation}
This section designs an exact magnetic field approximation using MLP that captures coil-specific geometric features in (\ref{CGL_circulant_integration_term}) along with its certification. 
The resulting pseudocode for offline training and online inference is presented in Algorithm~\ref{CGL_pseudocode_of_coil_geometry_learning}.
\subsection{Minimum-Dimensional Sample Collection and Learning}
\label{CGL_Reduced_Order_Sample_Collection_and_Learning}
We first introduce the data collection strategy for consistent learning and reducing the sample space. For arbitrary $j$th and $k$th agents with the coil radius $a_{\mathrm{NN}}$, we define arbitrary attitude parameters $\sigma_j$ and $\sigma_k$ from the arbitrary reference frame $\mathcal{A}$ to $j$th and $k$th body-fixed frame, respectively, and assume the coil geometry information ${{g}}^a_{j_{v}\leftarrow k_w}$ in (\ref{CGL_circulant_integration_term}) is given:
$$
{{g}}^a_{j_{v}\leftarrow k_w}(r^{a}_{j\leftarrow k},\sigma_j,\sigma_k,a_{\mathrm{NN}})\in\mathbb{R}^6.
$$
We reduce its sample region from 
$(r^{a}_{j\leftarrow k},\sigma_j,\sigma_k)\in\mathbb{R}^9$ to 
$(\hat{r}_{j\leftarrow k}^{d_{k_{w}}},\phi)\in\mathbb{R}^4$ by the two-coordinate transformation: 1) the $k$th reference coil lies within the $x–y$ plane and 2) the $j$th coil lies within the $x_+–z_+$ plane of the $k$th reference coil, as illustrated in Fig.~\ref{CGL_fig:coil_geometry_learning}. We first define the unit vectors $\bm n_{k(x,y,z)}$ along with each axes of $k$th body-fixed frame $C^{A/B_k}(\sigma_k)$:
\begin{equation*}
\begin{bmatrix}
    n^a_{k_{x}}&n^a_{k_{y}}&n^a_{k_{z}}
\end{bmatrix}\triangleq C^{A/B_k}(\sigma_k).
\end{equation*}
We define three coordinate frames $B_{k_{x}},B_{k_{y}},B_{k_{z}}$ such that each $n_{k(x,y,z)}$ corresponds to its z-axis and their transformation matrices $C^{A/B_{k(x,y,z)}}\in\mathbb{R}^{3\times 3}$ for three axes are
\begin{equation}
\label{CGL_Bjxyz2A}
C^{A/B_{k_{\mathbf{(z,x,y)}}}}\triangleq
\begin{bmatrix}
n^a_{k_{(x,y,z)}}&n^a_{k_{(y,z,x)}}&n^a_{k_{\mathbf{(z,x,y)}}}
\end{bmatrix}
\end{equation}
where $C^{A/B_k}(\sigma_k)=C^{A/B_{k_{z}}}$ and this achieves that the $k_w$th reference coil lies within the $x–y$ plane. We next adjust the direction of $r_{j\leftarrow k}$ to make its $z$-component positive by
\begin{equation}
\label{CGL_first_coordinate_transformation}
\begin{aligned}
    \begin{bmatrix}
r^{b_{k_{w}}}_{j\leftarrow k}\\
        n^{b_{k_{w}}}_{j\leftarrow k}
    \end{bmatrix}
    \triangleq
    \left(
    I_2\otimes C^{B_{k_{w}}/A}\right)
    \begin{bmatrix}
\text{sgn}\left(n^{a\top}_{k_{w}} r^a_{j\leftarrow k}\right)\ r^a_{j\leftarrow k}\\
        n^a_{j\leftarrow k}
    \end{bmatrix}
\end{aligned}
\end{equation}
where $\text{sgn}(\cdot)$ denotes the sign function and $\text{sgn}(0)\triangleq 1$. 
The following transformation achieves that the $j$th coil lies within the $x_+–z_+$ plane of the $k_w$th reference coil:
\begin{equation}
\label{CGL_Bjxyz2D}
\begin{aligned}
C^{D_{j\leftarrow k_w}/A}&=C^{D_{j\leftarrow k_w}/B_{k_w}}C^{B_{k_w}/A}\\
C^{D_{j\leftarrow k_{w}}/B_{k_{w}}}&\triangleq
\begin{bmatrix}
c_{\theta_{b_{k_{w}}}}&s_{\theta_{b_{k_{w}}}}&0\\
-s_{\theta_{b_{k_{w}}}}&c_{\theta_{b_{k_{w}}}}&0\\
0&0&1
\end{bmatrix}
\end{aligned}
\end{equation}
where $\theta_{b_{k_{w}}}\triangleq \tan^{-1}{(r^{b_{k_{w}}}_{j\leftarrow k(2)},r^{b_{k_{w}}}_{j\leftarrow k(1)})}$. Since this imposes $r^{d_{k_{w}}}_{j\leftarrow k(2)}=0$, 
we reduce the input into 
$\hat{r}^{d_{k_{w}}}_{j\leftarrow k}\in\mathbb{R}^2$. Moreover, 
$n^{d_{k_{w}}}_{j_{v}}=[c_{\phi_{(1)}}s_{\phi_{(2)}};s_{\phi_{(1)}}s_{\phi_{(2)}};c_{\phi_{(2)}}]\in\mathbb{R}^3$ is reduced into $\phi\in\mathbb{R}^2$ by the spherical coordinate representation using azimuth $\phi_{(1)}\in(-\pi,\pi]$ and elevation $\phi_{(2)}\in(0,\pi)$. As a result, we obtain the reduced-dimensional input vector: 
\begin{equation}
\hat{r}^{d_{k_{w}}}_{j\leftarrow k}\triangleq
\begin{bmatrix}    
r^{d_{k_{w}}}_{j\leftarrow k(1)}\\
r^{d_{k_{w}}}_{j\leftarrow k(3)}
\end{bmatrix},\quad \phi \triangleq \begin{bmatrix} \tan^{-1}\left(n^{d_{k_{w}}}_{j_{v}(2)},n^{d_{k_{w}}}_{j_{v}(1)}\right)\\
\cos^{-1}\left(n^{d_{k_{w}}}_{j_{v}(3)}\right)
\end{bmatrix}
\end{equation}
To compensate the sign change of input data in (\ref{CGL_first_coordinate_transformation}), the label data is also modified as follows:
\begin{equation}
\label{CGL_coil_geometry_vector_dkw}
    {g}^{d_{k_{w}}}_{j_{v}\leftarrow k_w}
\triangleq\left(
\begin{bmatrix}
    \text{sgn}\left(n^{a\top}_{k_{w}} r^a_{j\leftarrow k}\right)&0\\
    0&1
\end{bmatrix}
\otimes C^{D_{j\leftarrow k_{w}}/A}\right)
    {g}^a_{j_{v}\leftarrow k_w}
\end{equation}
These derive the input $(\hat{r}_{j\leftarrow k}^{d_{k_{w}}},\phi)$ and label ${g}^{d_{k_{w}}}_{j_{v}\leftarrow k_w}$ and we stack $N_s$ data $X=[\boldsymbol{r},\boldsymbol{\phi}]\in\mathbb{R}^{N_s\times 4}$ and $Y\in\mathbb{R}^{N_s\times 6}$
\begin{equation*}
\boldsymbol{r}=
\begin{bmatrix}
\hat{r}_{j\leftarrow k}^{{d_{k_{w}}}(1)\top}\\
\vdots \\
\hat{r}_{j\leftarrow k}^{{d_{k_{w}}}(N_s)\top}
\end{bmatrix}
,\boldsymbol{\phi}=
\begin{bmatrix}
\phi^{(1)\top}\\
\vdots \\
\phi^{(N_s)\top}
\end{bmatrix}
,\boldsymbol{Y}=
\begin{bmatrix}
g_{j_z\leftarrow k_z}^{{d_{k_{w}}}(1)\top}
\\
\vdots \\
g_{j_z\leftarrow k_z}^{{d_{k_{w}}}(N_s)\top}
\end{bmatrix}
\end{equation*}
We train our MLP $\mathfrak{g}$ to minimize an arbitrary loss function $f(\cdot)$: $\theta_{w}^*,\theta_{b}^*= \mathrm{argmin}\sum_{j=1}^{N_s}f(\|\bm{Y}^{(j)\top}-\hat{\bm{Y}}^{(j)\top}\|)$ where $\hat{\bm{Y}}^{(j)}$ is estimated values by $\mathfrak{g}$ using training data set $\mathcal{T}_{a_{\mathrm{NN}}}=\{(X_{}^{(i)},Y_{}^{(i)})\}_{i=1}^N$ in given compact set $\mathcal{S}_{a_{\mathrm{NN}}}$ with $2a_{\mathrm{NN}}<\underline{r}$:  
$$
\mathcal{S}_{a_{\mathrm{NN}}}=\left\{
\begin{aligned}
r_{j\leftarrow k}&\in\mathbb{R}^2\\    
\phi&\in\mathbb{R}^2
\end{aligned}
\ \left|\ \underline{r} \leq \|r_{j\leftarrow k}\|\leq \overline{r},\quad 
\begin{aligned}
&\phi_{(1)}\in[-\pi,\pi]\\ 
&\phi_{(2)}\in[0,\pi]
\end{aligned}
\right\}\right.
$$
Here, we obtain $\hat{\bm{Y}}^{(j)}=W_{dl}\times\mathfrak{g}{\left(W_{di}^{-1}(X^{(j)}-b_{di}), \bm{\theta}\right)}+b_{dl}$ by $\mathfrak{g}$ using preprocessing matrices $W_{dl,di}$ and vectors $b_{dl,di}$, e.g., normalization or standardization. 
\subsection{Generalization for Arbitrary Radius Coil Inference}
We can apply the learned model $\mathfrak{g}$ for coil radius $a_{\mathrm{NN}}$ to different coils with coil radius $a_{\mathrm{inf}}\triangleq\gamma a_{\mathrm{NN}}$ via appropriate transformations without retraining. 
\begin{theorem}
\label{CGL_Generalization_CGL}
For given MLP model $\mathfrak{g}{(\bm{X}, \bm{\theta})}=\mathfrak{g}{\left([\bm{r},\bm{\phi}], \bm{\theta}\right)}$ learned coil geometry information for coil radius $a_{\mathrm{NN}}$ in $\mathcal{S}_{a_{\mathrm{NN}}}$, consider the coil radius $a_{\mathrm{inf}}\triangleq \gamma a_{\mathrm{NN}}$ and 
the new data 
in the sample region $\mathcal{S}_{a_{\mathrm{inf}}}$ that is bounded by the constants of $\mathcal{S}_{a_{\mathrm{NN}}}$:
$$
\mathcal{S}_{a_{\mathrm{inf}}}=\left\{
\begin{aligned}
{r}_{j\leftarrow k}&\in\mathbb{R}^2\\    
\phi&\in\mathbb{R}^2
\end{aligned}
\ \left|\ 
\underline{r} \leq \frac{\|{r}_{j\leftarrow k}\|}{\gamma}\leq \overline{r},\quad 
\begin{aligned}
&\phi_{(1)}\in[-\pi,\pi]\\ 
&\phi_{(2)}\in[0,\pi]
\end{aligned}
\right\}\right..
$$
Then, $\mathfrak{g}$ derives ${{g}}^a_{j_{v}\leftarrow k_w}({r}^{a}_{j\leftarrow k},\sigma_j,\sigma_k,a_{\mathrm{inf}})$ as 
\begin{equation}
\label{CGL_mutual_relationship}
{{g}}^a_{j_{v}\leftarrow k_w}({r}^{a}_{j\leftarrow k},\sigma_{j},\sigma_k,a_{\mathrm{inf}})
=\begin{bmatrix}
    I_3&O\\
    O&\gamma I_3
\end{bmatrix}\mathfrak{g}{\left(\left[\frac{{{r}}^{(i)}}{\gamma},\bm{\phi}^{(i)}\right], {\theta}\right)}
\end{equation}
\end{theorem}
\begin{proof}
Substituting ${{r}}/{\gamma}$ into 
${{g}}^a_{j_{v}\leftarrow k_w}$ in (\ref{CGL_circulant_integration_term}) yields 
(\ref{CGL_mutual_relationship}).
\end{proof}
\begin{remark}
\textcolor{black}{The universality of the proposed framework, as established in Theorem~\ref{CGL_Generalization_CGL}, is validated in section~\ref{CGL_Numerical_Validation} through a spacecraft docking scenario.  
The model is trained only once with $a_{\mathrm{NN}}=0.3$ in subsection~\ref{CGL_Offline_Training} and is directly deployed—via transformations in (\ref{CGL_mutual_relationship}) without retraining—to numerical simulations with $a_{\mathrm{inf}}=0.15$ in subsection~\ref{CGL_numerical_validation} and experimental tests with $a_{\mathrm{inf}}=0.075$ in subsection~\ref{CGL_experimental_validation}.}
\end{remark}
\subsection{Certified Error Bounds of Coil Geometry Information}
Finally, we derive the theoretical upper bound on the learning error. Followed by (\ref{CGL_upper_bound_learning_error}), we obtain the error bound 
\begin{equation}
\label{CGL_prediction_error_lipchitz}
\sup_{x\in \mathcal{S}_{a_{\mathrm{NN}}}}
\Delta_{g(x)}
\leq 
\sup_{x'\in \mathcal{T}_{a_{\mathrm{NN}}}}
\Delta_{g(x')}
+ 
({L_g}+{L_\mathfrak{g}})
\begin{bmatrix}
\rho_{\mathcal{S}_{(a_{\mathrm{NN}},r)}}\\
{\rho_{\mathcal{S}_{(a_{\mathrm{NN}},\phi)}}}
\end{bmatrix}    
\end{equation}
where a constant Lipchitz matrix of true function $L_{{g}}\in\mathbb{R}^{2\times 2}$ and MLP $L_{\mathfrak{g}}\in\mathbb{R}^{2\times 2}$, respectively, and covering radius about position and attitude $\rho_{\mathcal{S}_{\{(a_{\mathrm{NN}},r),(a_{\mathrm{NN}},\phi)\}}}$. We can derive 
${L_\mathfrak{g}}$ in (\ref{CGL_lipchitz_constant_MLP}) as in subsection~\ref{CGL_certified_error_bound} and
the next theorem gives $L_{{g}}\in\mathbb{R}^{2\times 2}$ and $\rho_{\mathcal{S}_{\{(a_{\mathrm{NN}},r),(a_{\mathrm{NN}},\phi)\}}}$. \textcolor{black}{This $L_{g}$ and the error bound in (\ref{CGL_upper_bound_learning_error}) provide a guideline for sampling density.}
\begin{theorem}[Certified Error Bounds]
\label{CGL_theorem_Certified_Error_Bounds}
Consider $N$ optimally placed training data set $\mathcal{T}_{a_{\mathrm{NN}}}$ in $S_{a_{\mathrm{NN}}}$. Then, the prediction error for a coil radius $a_{\mathrm{NN}}$ is bounded as (\ref{CGL_prediction_error_lipchitz}) where $L_0={2(2\pi a_{\mathrm{NN}})^2}/{r_{\min}^3}$, $r_{\min}\triangleq\inf_{\theta_{j_v},\theta_{k_w}}\|{r}_{j_{v}\leftarrow k_w}\|$,
\begin{equation}
\label{CGL_eq_certified_error_bound}
\begin{aligned}
&\left(L_g+L_{\mathfrak{g}}\right)
    \begin{bmatrix}
\rho_{\mathcal{S}_{(a_{\mathrm{NN}},r)}}\\
{\rho_{\mathcal{S}_{(a_{\mathrm{NN}},\phi)}}}
    \end{bmatrix}\\
=&\left(L_0
\begin{bmatrix}
1&a_{\mathrm{NN}}+r_{\mathrm{min}}/2\\
a_{\mathrm{NN}}&a_{\mathrm{NN}}(a_{\mathrm{NN}}+r_{\min})
    \end{bmatrix}+L_{\mathfrak{g}}\right)
    \begin{bmatrix}
        \frac{\sqrt{\overline{r}^2-\underline{r}^2}}{2N^{1/4}}\\
        \frac{\sqrt{2\pi}}{N^{1/4}}
    \end{bmatrix}
\end{aligned}.
\end{equation}
The prediction error bound for a coil radius $a_{\mathrm{inf}}\triangleq \gamma a_{\mathrm{NN}}$ is
\begin{equation}
\label{CGL_prediction_error_lipchitz_different_coil}
\sup_{x\in \mathcal{S}_{a_{\mathrm{inf}}}}
\Delta_{g(x)}=
\begin{bmatrix}
    I_3&O\\
    O&\gamma I_3
\end{bmatrix}\sup_{x\in \mathcal{S}_{a_{\mathrm{NN}}}}\Delta_{g(x)}.
\end{equation}
\end{theorem}
\begin{proof}
    See Appendix~\ref{CGL_proof_Certified_Error_Bounds}.
\end{proof}
\begin{algorithm}[tb]
\caption{Coil geometry learning in reference frame ($A$)}
\begin{algorithmic}[1]
\STATE \textbf{--- Offline Phase ---------------------------------------------}
\STATE \textbf{Inputs: }sample number $n_{\mathrm{s}}$
\STATE\textbf{Outputs: }MLP weights $\theta_{w}^*$, MLP bias $\theta_{b}^*$
\FOR{$i\in[1,\dots, n_{\mathrm{s}}]$}
  \STATE Generate random $r_{j\leftarrow k}^{a(i)}$ and $n_{{j,k}_{(x,y,z)}}^{a(i)}$ 
  \STATE Derive $C^{A/B_{k_z}}=
\begin{bmatrix}
    n_{k_x}^a&n_{k_y}^a&{^a}n_{k_z}^a
\end{bmatrix}$ in (\ref{CGL_Bjxyz2A})
  \STATE Derive $C^{D_{j\leftarrow {k_z}}/B_{k_z}}$ in (\ref{CGL_Bjxyz2D}) and $r_{j\leftarrow k}^{d_{j\leftarrow k_w}}$
  \STATE Stack inputs $\bm{X}^{(i)}=[r_{j\leftarrow k}^{d_{j\leftarrow {k_z}}(i)};n_{j_z}^{d_{j\leftarrow {k_z}}(i)}]^\top$
  \STATE Calculate labels $\bm{Y}^{(i)}={g}^{d_{j_{v}\leftarrow {k_w}}}_{j_{v}\leftarrow {k_w}}$ in (\ref{CGL_coil_geometry_vector_dkw})
\ENDFOR
\STATE Derive $\theta_{w}^*,\theta_{b}^*$ by $\bm{X}$ and $\bm{Y}$, e.g., (\ref{CGL_huber_learning_function})
\STATE Derive $\sup_{x\in \mathcal{S}_{a_{\mathrm{inf}}}}
\Delta_{g(x)}$ in Theorem~\ref{CGL_theorem_Certified_Error_Bounds} where ${L_\mathfrak{g}}\triangleq \|\phi\|_{\text {Lip }}^L\prod_{l=2}^{L} \sigma(W^{(l)})[L_{\mathfrak{g}_{\mathrm{pos}}}^{\mathrm{pos}},L_{\mathfrak{g}_{\mathrm{pos}}}^{\mathrm{rot}};L_{\mathfrak{g}_{\mathrm{rot}}}^{\mathrm{pos}},L_{\mathfrak{g}_{\mathrm{rot}}}^{\mathrm{rot}}]$ and 
\begin{equation}
\label{CGL_lipchitz_constant_MLP}
{\small
\left\{
\begin{aligned}
L_{\mathfrak{g}_{\mathrm{pos}}}^{\mathrm{pos}}&=
\sigma\left(W_{dl(1:3,:)}W^{(L+1)}\right)\ \sigma\left(W^{(1)}W_{di(:,1:2)}^{-1}\right)\\
L_{\mathfrak{g}_{\mathrm{pos}}}^{\mathrm{rot}}&=
\sigma\left(W_{dl(1:3,:)}W^{(L+1)}\right)\ \sigma\left(W^{(1)}W_{di(:,3:4)}^{-1}\right)\\
L_{\mathfrak{g}_{\mathrm{rot}}}^{\mathrm{pos}}&=
\sigma\left(W_{dl(4:6,:)}W^{(L+1)}\right)\ \sigma\left(W^{(1)}W_{di(:,1:2)}^{-1}\right)\\
L_{\mathfrak{g}_{\mathrm{rot}}}^{\mathrm{rot}}&=
\sigma\left(W_{dl(4:6,:)}W^{(L+1)}\right)\ \sigma\left(W^{(1)}W_{di(:,3:4)}^{-1}\right)\\
\end{aligned}
\right.
}
\end{equation}
\STATE \textbf{--- Online Phase -----------------------------------------------}
\STATE \textbf{Inputs: }1) $r^{a}_{j\leftarrow k}$, 2) $n^{a}_{{j,k}_{(x,y,z)}}$, 3) $\theta_{w}^*,\ \theta_{b}^*$, 4) $A$
\STATE\textbf{Outputs: }Exact model coefficient matrix $Q^a_{j\leftarrow k}\in\mathbb{R}^{6\times 9}$
\FOR{$k_w\in[k_x,k_y,k_z]$}
\STATE Derive $C^{D_{j\leftarrow k_w}/A}$ in (\ref{CGL_Bjxyz2D}) and $r_{j\leftarrow k}^{d_{j\leftarrow k_w}}$
\FOR{$j_{v}\in[j_x,j_y,j_z]$}
\STATE Derive $n_{j_{v}}^{d_{j\leftarrow k_w}}$ and stack $x=\left[r_{j\leftarrow k}^{d_{j\leftarrow k_w}\top},n_{j_{v}}^{d_{j\leftarrow k_w}\top}\right]$
\STATE Infer ${g}^{d_{j_{v}\leftarrow {k_w}}}_{j_{v}\leftarrow {k_w}}=\mathfrak{g}(x,\bm\theta)$ in (\ref{CGL_coil_geometry_vector_dkw})
\STATE Calculate $g_{j_{v}\leftarrow {k_w}}^a= (I_2\otimes C^{A/D_{j\leftarrow k_w}}){g}^{d_{j_{v}\leftarrow {k_w}}}_{j_{v}\leftarrow {k_w}}$
\STATE Calculate 
$g^a_{j_{v}\leftarrow k_w\mathrm{off}}=
\begin{bmatrix}
    I_3&O_3\\
\left[r_{j_{v}\mathrm{off}}\right]_\times&I_3
\end{bmatrix}
{{g}}^a_{j_{v}\leftarrow k_w}$
\ENDFOR
\STATE Stack $[{{g}}_{j\leftarrow k_w}]=
\begin{bmatrix}
{{g}}_{j_{x}\leftarrow k_w\mathrm{off}}&
{{g}}_{j_{y}\leftarrow k_w\mathrm{off}}&
{{g}}_{j_{z}\leftarrow k_w\mathrm{off}}
\end{bmatrix}$
\ENDFOR
\STATE Stack ${{G}}_{j\leftarrow k}=A^{-2}
\begin{bmatrix}
[{{g}}_{j\leftarrow k_x}]&[{{g}}_{j\leftarrow k_y}]&[{{g}}_{j\leftarrow k_z}]
\end{bmatrix}$
\end{algorithmic}
\label{CGL_pseudocode_of_coil_geometry_learning}
\end{algorithm}
\section{Validation by Satellite Docking Application}
\label{CGL_Numerical_Validation}
We validate the proposed learning-based current calculation for the docking control of the two satellites. We primarily compare docking results obtained by the learned exact model and the far-field approximation model in subsection~\ref{CGL_appendix_far_field_approximation}. 
\subsection{Offline Training of Coil Geometry Model}
\label{CGL_Offline_Training}
\begin{table}[bt!]
\centering
\caption{Sample region information and covering radii.
}
\begin{tabular}{cccccc}
\hline 
&$\frac{\underline{r}/\overline{r}}{2a_{\mathrm{NN}}}$ (m)& N (-)&$\rho_{\mathcal{S}_{(a_{\mathrm{NN}},r)}}/\rho_{\mathcal{S}_{(a_{\mathrm{NN}},\phi)}}$\\ \hline 
$\mathcal{S}^A_{a_{\mathrm{NN}}}$& 1.0005 / 1.075& 1.0e$^9$&0.66e$^{-3}$ / 14.10e$^{-3}$& \\ 
$\mathcal{S}^B_{a_{\mathrm{NN}}}$& 1.4 / 1.9& 2e$^8$&3.24e$^{-3}$ / 21.08e$^{-3}$& \\ 
$\mathcal{S}^C_{a_{\mathrm{NN}}}$& 1.046 / 4.0 & 2.5e$^7$&16.38e$^{-3}$ / 35.45e$^{-3}$& \\
\hline
\end{tabular}
\end{table} 
We first train the magnetic interaction model $\mathfrak{g}$ by coil radius $a_{\mathrm{NN}}=0.3$ based on our framework. This study collects 11 million samples in three domains $\mathcal{S}^A_{a_{\mathrm{NN}}},\mathcal{S}^B_{a_{\mathrm{NN}}},\mathcal{S}^C_{a_{\mathrm{NN}}}$ as shown in Table~\ref{CGL_Offline_Training}. \textcolor{black}{After model tuning process to minimize network size}, we use a model with two hidden layers, each with 256 and 128 neurons, respectively, that use Layer Normalization and Gaussian Error Linear Unit activation \cite{hendrycks2016gaussian} whose Lipschitz constant is bounded $\|\phi\|_{\text{Lip}}
\leq 1.129$. We train by the Adam optimizer \cite{kingma2014adam}, a batch size of 131,072 for pre-training and 1,024 for fine-tuning, and the smooth L1 loss function $f(\cdot)$, with a spectral normalization regularization term \cite{shi2019neural}:
\begin{equation}
\label{CGL_huber_learning_function}
\begin{aligned}
\mathcal{L}(\theta_w,\theta_b)
&=\sum_{j=1}^{N_s}
\sum_{i=1}^{6}
{\rho_\delta\left(
Y_i^{(j)} - \hat{Y}_i^{(j)}
\right)}/{(6N_s)}\\
&+
\lambda_{\mathrm{sn}}
\sum_{l=1}^{L+1}
\left[
\max\left(0,\,\sigma_{\max}({W}^{(l)})-\overline{w}\right)
\right]^2
\end{aligned}
\end{equation}
where Huber loss function $\rho_\delta(r_i) =r_i^2/2$ if $|r_i| < \delta$ and otherwise, $\delta(|r_i|-{1}/{2})$, and $\overline{w}$ is the user-defined constant. An implementation that enforces the Lipschitz constant exactly, rather than using a penalty, is described in \cite{miyato2018spectral,shi2019neural}. The learning rate is scheduled using cosine annealing from $1e^{-2}$ to $1e^{-6}$ for pre-training and $1e^{-6}$ to $1e^{-\textcolor{black}{9}}$ for fine-tuning.

The training process for $\lambda_{\mathrm{sn}}=0$ converged with a test loss of $4.11 e^{-3}$, a training loss of $3.01 e^{-3}$, and $\overline{\sigma}(L_\mathfrak{g})=1.17e^{7}$ by $L_{\mathfrak{g}}\lessapprox \sigma(W_{dl})(\|\phi\|_{\text {Lip }}^L\prod_{l=1}^{L+1} \sigma(W^{(l)}))\sigma\left(W_{di}^{-1}\right)$. We calculate the Lipschitz constant of a true function as $\overline{\sigma}(L_{g})=40.47$ in $\mathcal{S}^A_{a_{\mathrm{NN}}}$ and $\overline{\sigma}(L_\mathfrak{g})$ is significantly larger than the true one. This motivates us to add a spectral normalization term. The training process for $\lambda_{\mathrm{sn}}=1e^{-3}$ converged with a test loss of $5.29 e^{-3}$, a training loss of $5.34 e^{-3}$, and $\overline{\sigma}(L_\mathfrak{g}^{\mathrm{SN}})=44.79$. This agreement between the two losses indicates acceptable generalization performance. 
\begin{remark}
\textcolor{black}{The bound in (\ref{CGL_upper_bound_learning_error}) is valid but conservative; thus we used a relatively large dataset of 11 million samples (batch size 131,072). Preliminary tests show comparable performance with significantly fewer samples, and tighter probabilistic bounds, e.g., \cite{bartlett2017spectrally}, could further support these improvements.
}    
\end{remark}
\subsection{Numerical Validation for Spacecraft Docking}
\label{CGL_numerical_validation}
\begin{figure}[tb!]
  \centering
  \begin{minipage}[t]{0.485\FigWidth}
    \centering    
    \subfloat[Far‐field model result at 250s.]{\includegraphics[width=1\columnwidth]{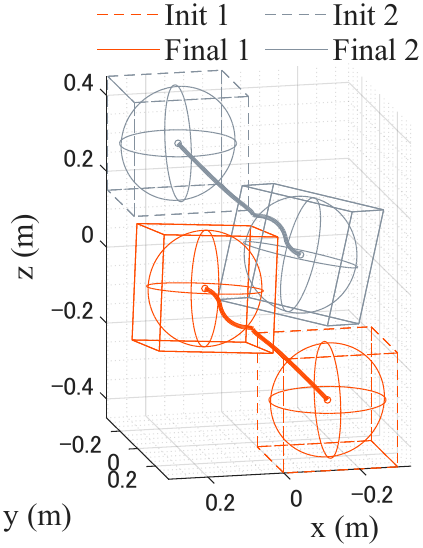}
    \label{CGL_fig:dock_far}}
  \end{minipage}%
  \begin{minipage}[t]{0.465\FigWidth}
    \centering
\subfloat[Exact model results at 300s.]{\includegraphics[width=1\columnwidth]{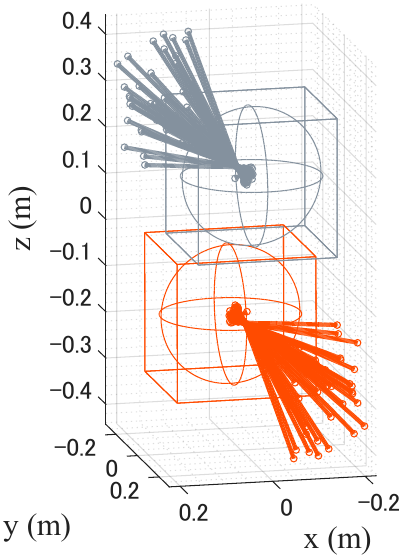}
    \label{CGL_fig:dock_near}}
  \end{minipage} 
  \begin{minipage}[t]{0.495\FigWidth}
    \centering
\subfloat[Position control results]{\includegraphics[angle=90,width=1.05\linewidth]{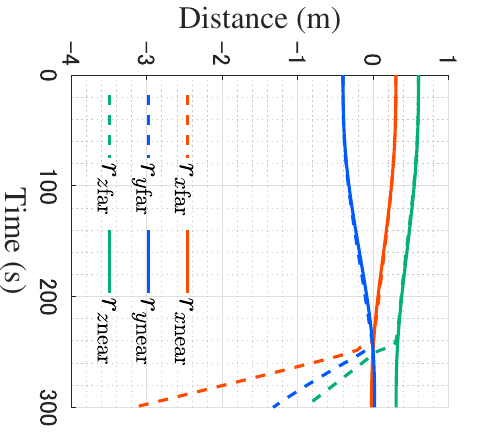}
    \label{CGL_fig:pos_control}}
  \end{minipage}%
  \begin{minipage}[t]{0.495\FigWidth}
    \centering
\subfloat[Force histories]{\includegraphics[angle=90,width=1.05\linewidth]{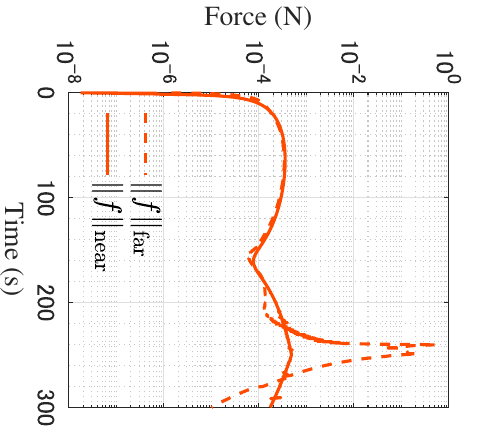}
    \label{CGL_fig:force_hist}}
  \end{minipage}
  \begin{minipage}[t]{0.495\FigWidth}
    \centering  
    \subfloat[Angular velocity control results]{\includegraphics[angle=90,width=1.05\linewidth]{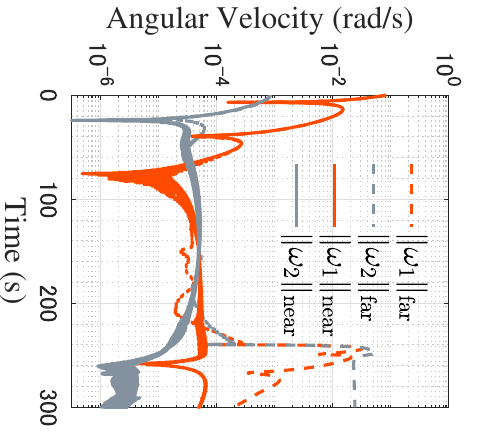}
    \label{CGL_fig:angvel_control}}
  \end{minipage}%
  \begin{minipage}[t]{0.495\FigWidth}
    \centering
\subfloat[Torque histories]{\includegraphics[angle=90,width=1.05\linewidth]{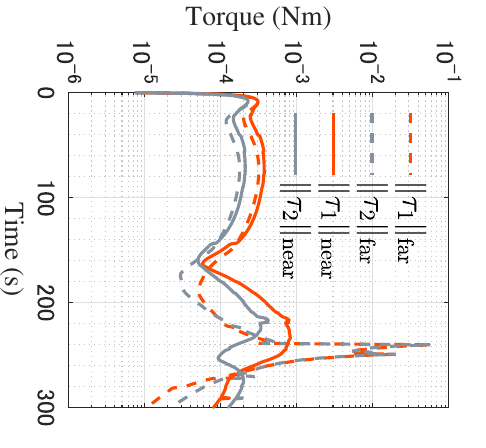}
    \label{CGL_fig:torque_hist}}
  \end{minipage}  
  \caption{Numerical results of two‐satellite docking control using the far‐field magnetic field approximation in Appendix~\ref{CGL_appendix_far_field_approximation} (Dashed lines) and the exact magnetic field model in (\ref{CGL_near_field_electromagnetic_interaction_model}) (solid lines). The target and chaser satellites are depicted in gray and red, respectively. 
    }
\label{CGL_Docking_distance_force_angular_torque}
\end{figure}

We first demonstrate the effectiveness of the learned exact model through numerical docking simulations. To achieve the target states, we design PD controllers and update the reference states using third-order splines at 0.1-second intervals to satisfy the current upper limit. \textcolor{black}{The detailed simulation and control settings including the controller gains and discretization method follow \cite{tajima2023study}.} 
Numerical integration uses the exact model in (\ref{CGL_near_field_electromagnetic_interaction_model}) and confirms angular momentum conservation, validating our simulation. For the dipole allocation to achieve controller input, we assume the following.
\begin{assumption}
\label{CGL_target_chaser_assumption}
Only the 1st (target) satellite uses reaction wheels (RWs) in Appendix~\ref{CGL_RW_attitude} for attitude control, whereas the 2nd (chaser) satellite uses electromagnetic torque.
\end{assumption}
\begin{assumption}
\label{CGL_fixed_current}
The current amplitudes on the 1st satellite are predetermined as $s^{b_1}_{1}=c_{\mathrm{amp}0}\eta(r_{12})[0.1,0.3,0.7]^\top$ and $c^{b_1}_{1}=c_{\mathrm{amp}0}\eta(r_{12})
    -s^{b_1}_{1}$ where $\eta(r_{12})={\|r_{12}(t)\|}/{\|r_{12}(0)\|}$ and 
    $c_{\mathrm{amp}0}=3$A to ensure regularity of inverse matrix in (\ref{CGL_decentralized_current_allocation}).
\end{assumption}
We compare the docking results using the exact model and the far-field approximation model. 
Figure~\ref{CGL_Docking_distance_force_angular_torque} shows the time evolution of the relative states and electromagnetic control inputs. For the far-field model, we observe significant input errors around 200-250 seconds, and this leads to a collision as shown in Fig.~\ref{CGL_fig:dock_far}. Conversely, Fig.~\ref{CGL_fig:dock_near} shows that the exact model accomplishes the docking control for randomly generated 30 initial states. It is also worth noting that the second satellite achieves attitude control solely through electromagnetic torque, which reaffirms the effectiveness of electromagnetic torque for attitude control \cite{takahashi2022kinematics,abbasi2022decentralized}.

\begin{figure}[tb!]
  \centering
  \subfloat[The time evolution of the distance between two satellites.]{%
    \includegraphics[width=1\FigWidth]{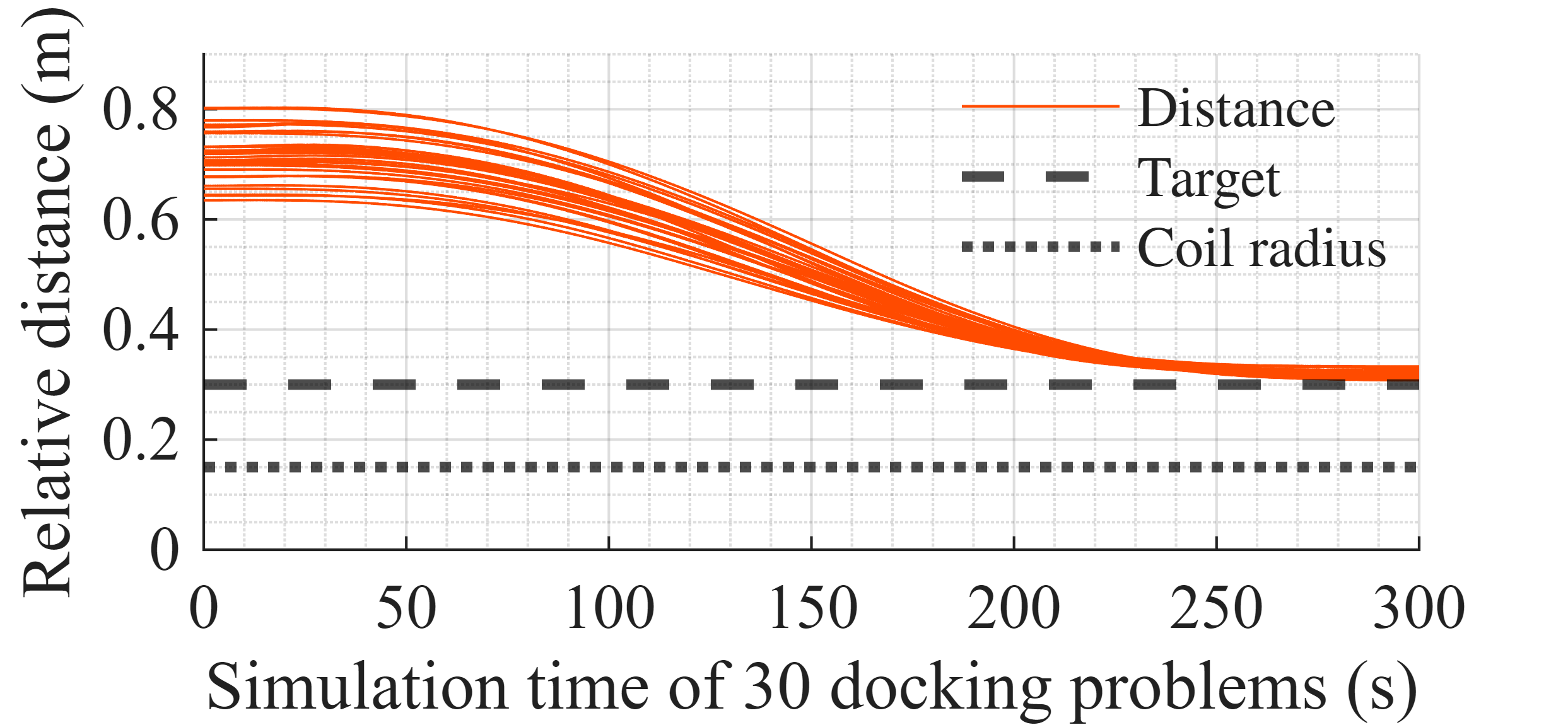}\label{CGL_fig:learning_control_distance}
  }\\
  \subfloat[Time comparison of direct computation and prediction in MATLAB R2025a with an AMD Ryzen Threadripper 7980X CPU (64 cores, 3.20 GHz base) and 64 GB RAM, Windows 11 Pro 64-bit. \label{CGL_fig:sample_region}]{\includegraphics[width=1\FigWidth]{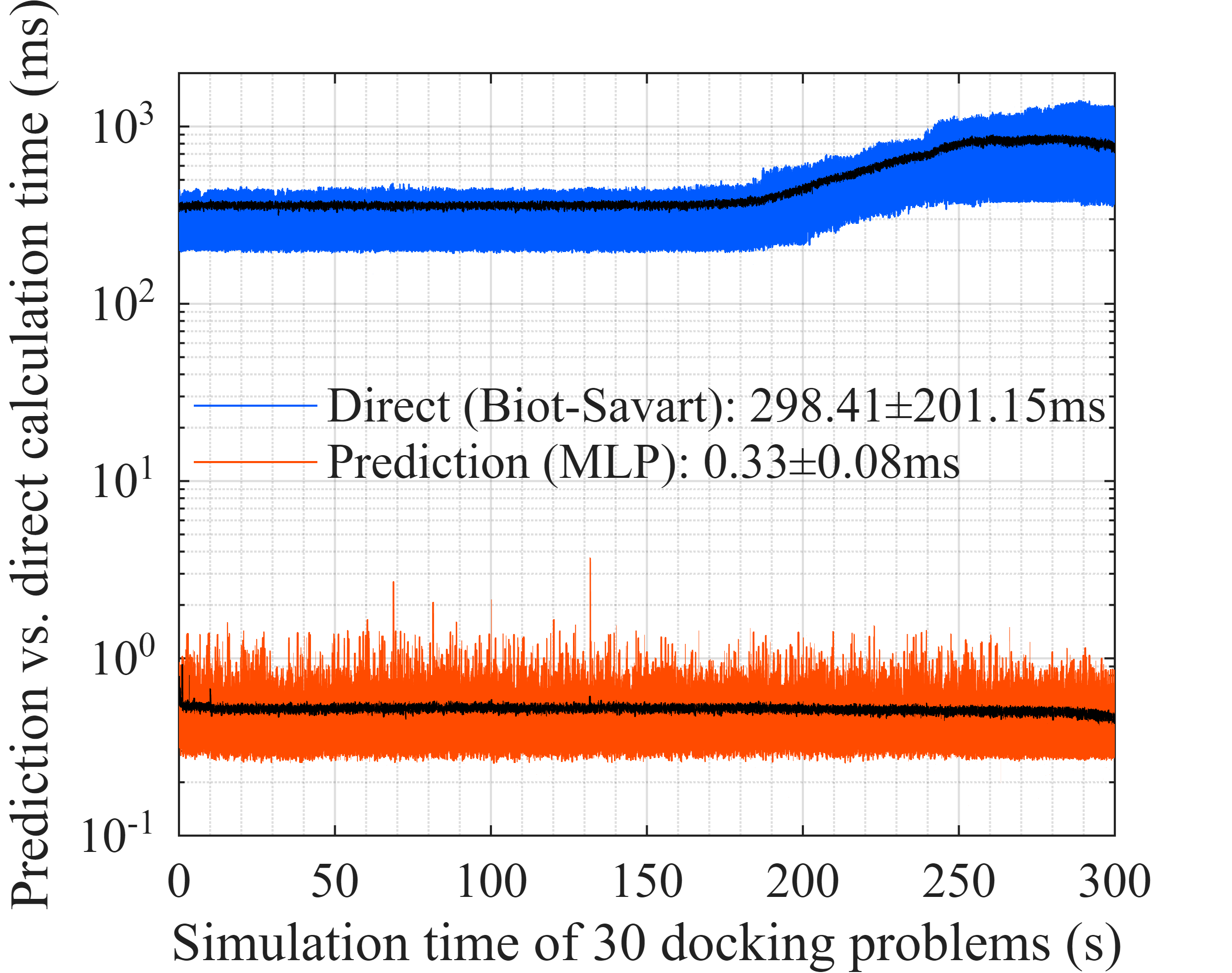}}\\
  \caption{The learned model-based docking control result. The time evolution of the distance and time comparison of direct computation (Biot–Savart double integral) and prediction using our model.}
\end{figure}

Next, we apply our trained magnetic interaction model $\mathfrak{g}$ to the docking control, and the results in Fig.~\ref{CGL_fig:learning_control_distance} show successful results for the newly generated 30 random initial states. On our PC, the average calculation time and its standard deviation decreased to approximately 1/1000 and 1/2500, respectively, indicating more consistent and stable computation. As the distance decreases, the Biot–Savart model increases the number of integration steps to ensure accuracy, whereas the MLP maintains a uniform computation regardless of the distance.
\subsection{Experimental Validation for Docking Control}
\label{CGL_experimental_validation}
\begin{figure}[!tb]
\centering
\begin{minipage}[b]{0.27\FigWidth}
\centering
\subfloat[Testbed \cite{takahashi2025noda_mmh}.]{\includegraphics[width=1\linewidth]{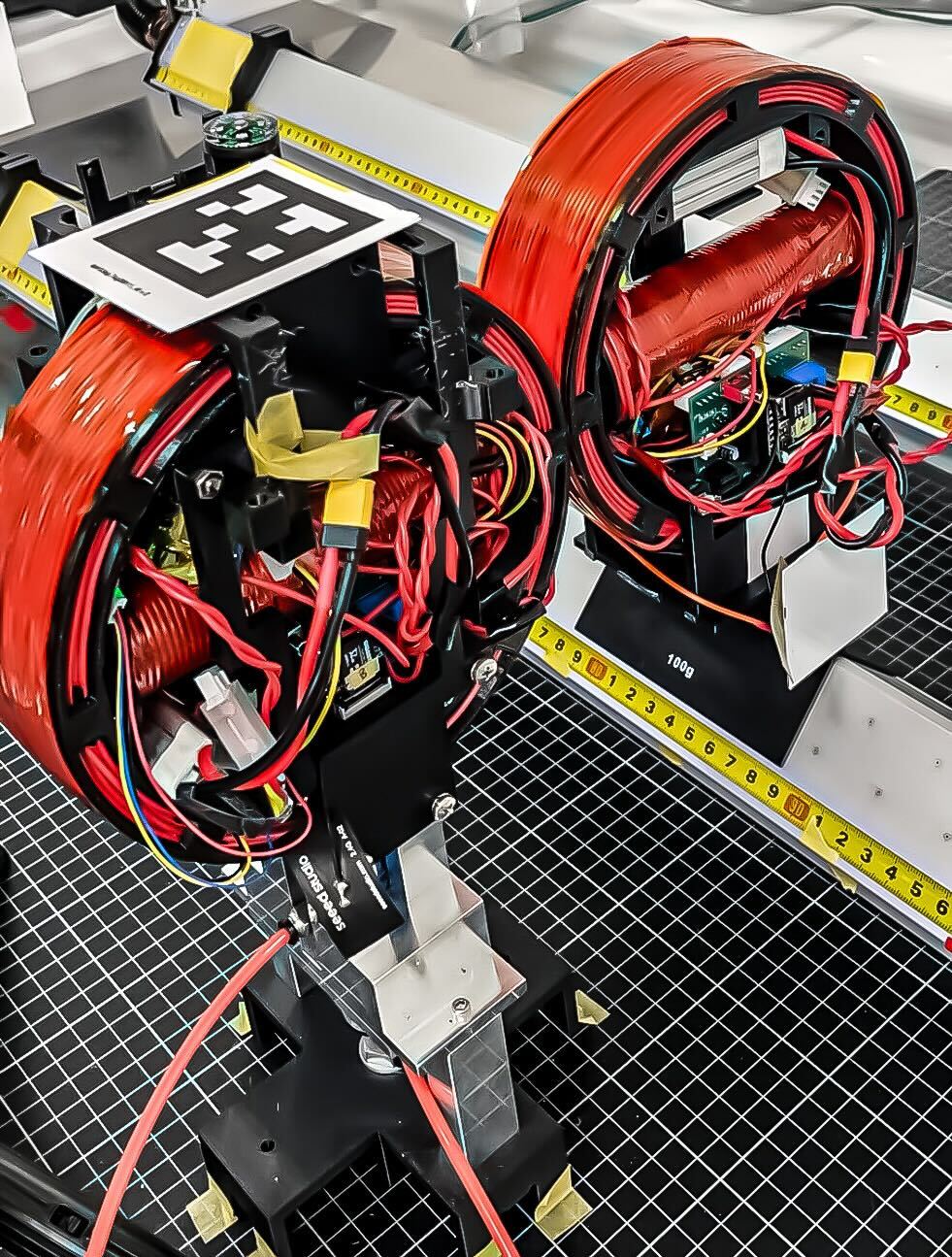}
\label{CGL_fig:setup}}
\end{minipage}
\hspace*{0.01\columnwidth}
\begin{minipage}[b]{0.325\FigWidth}
\centering
\subfloat[Initial condition.]{\includegraphics[width=1\linewidth]{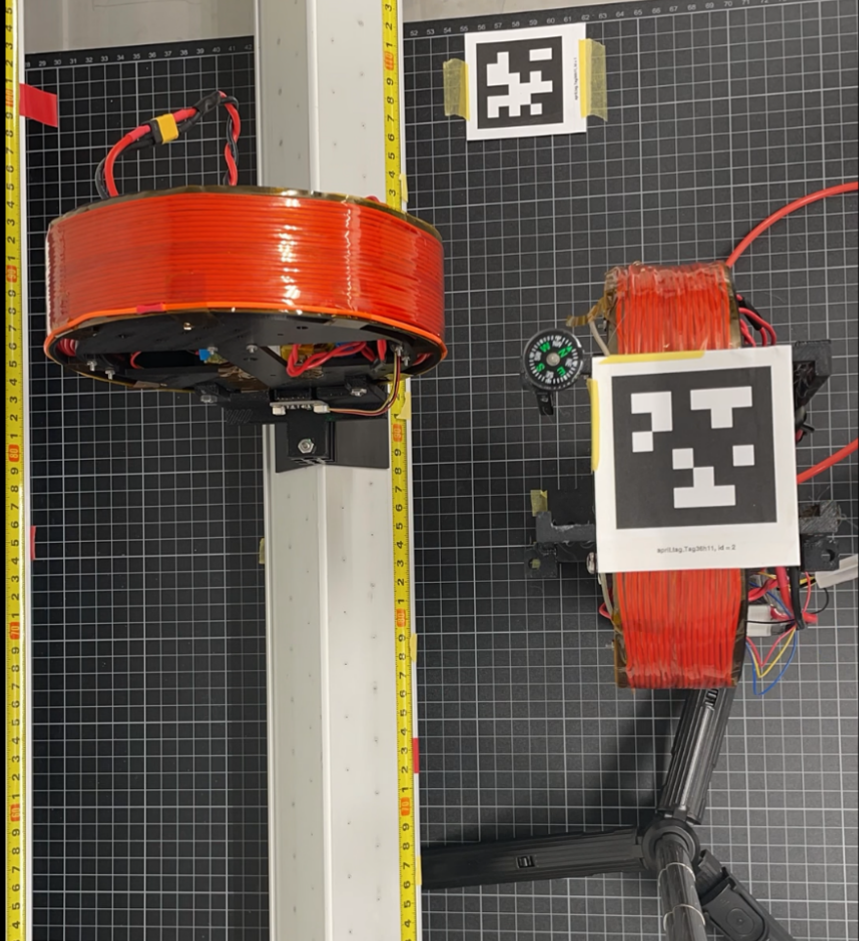}\label{CGL_fig:initial}}
\end{minipage}
\hspace*{0.01\columnwidth}
\begin{minipage}[b]{0.325\FigWidth}
\centering
\subfloat[Target condition.]{\includegraphics[width=1\linewidth]{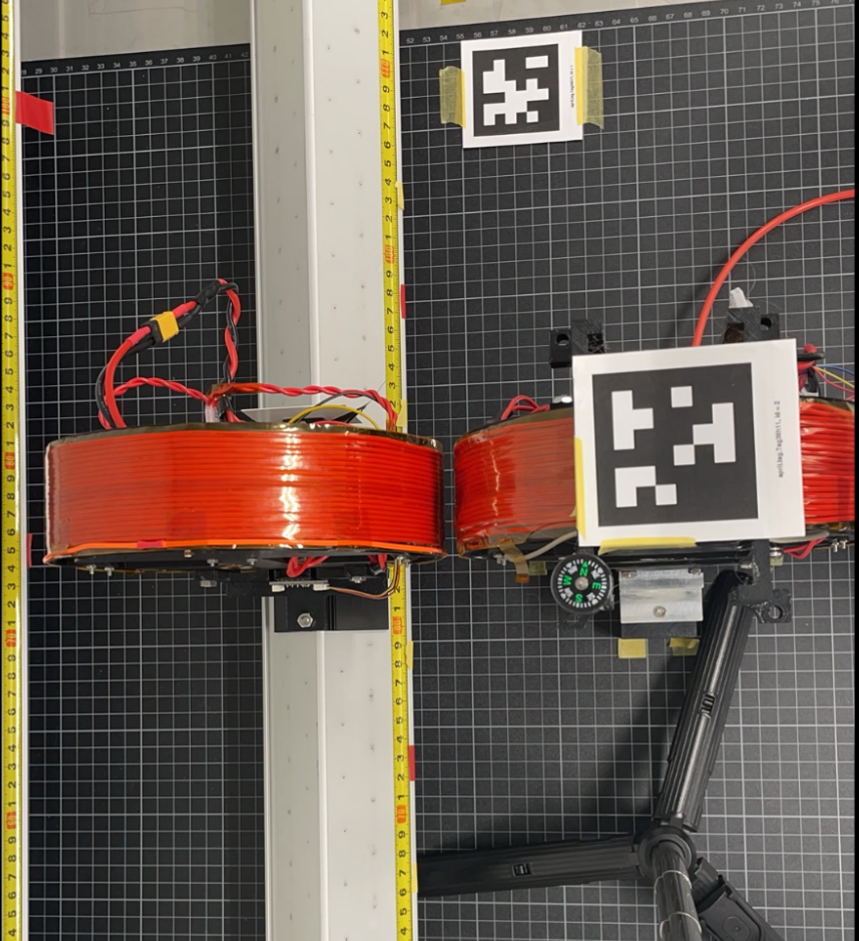}\label{CGL_fig:final}}
\end{minipage}
\begin{minipage}[b]{0.975\FigWidth}
\centering
\subfloat[Relative position control results.]{\includegraphics[width=1.0\linewidth]{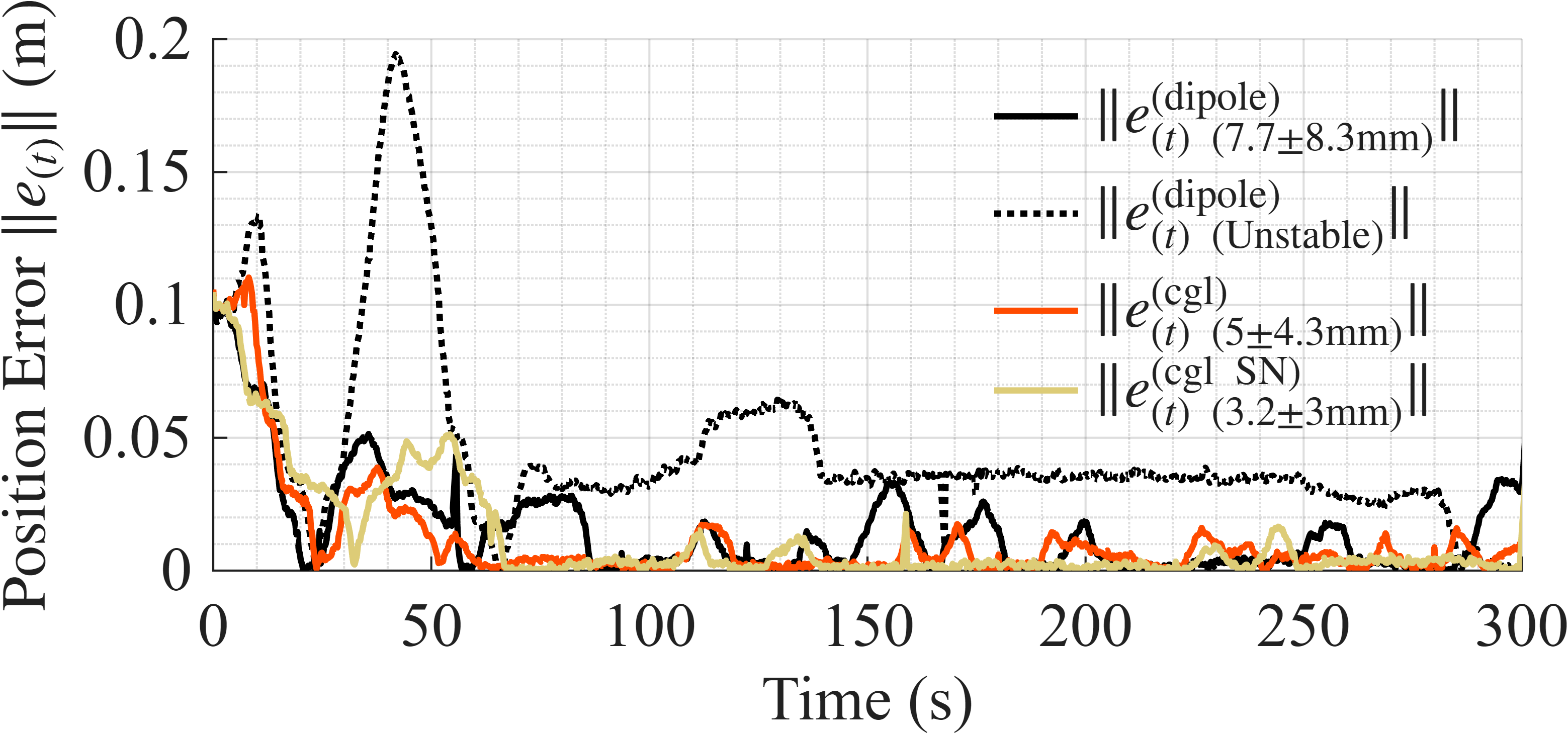}\label{CGL_fig:experimental_position_result}}
\end{minipage}
\begin{minipage}[b]{0.975\FigWidth}
\centering
\subfloat[Angle control results.]{\includegraphics[width=1\linewidth]{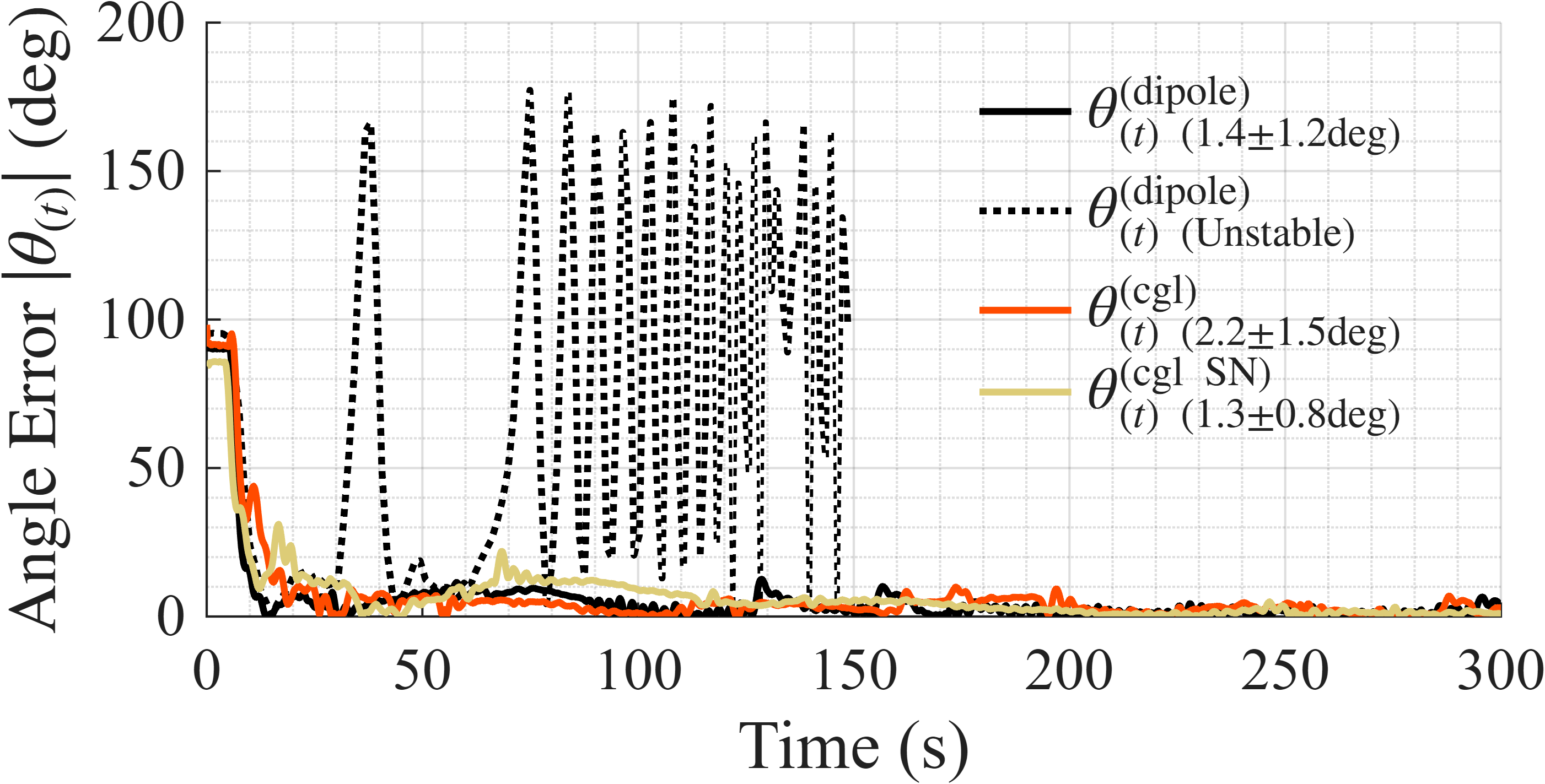}\label{CGL_fig:experimental_attitude_result}}
\end{minipage}\\
\begin{minipage}[b]{0.46\FigWidth}
\centering
\subfloat[Prediction error of $\|g^{\mathrm{pos}}\|$]{\includegraphics[width=1.05\linewidth]{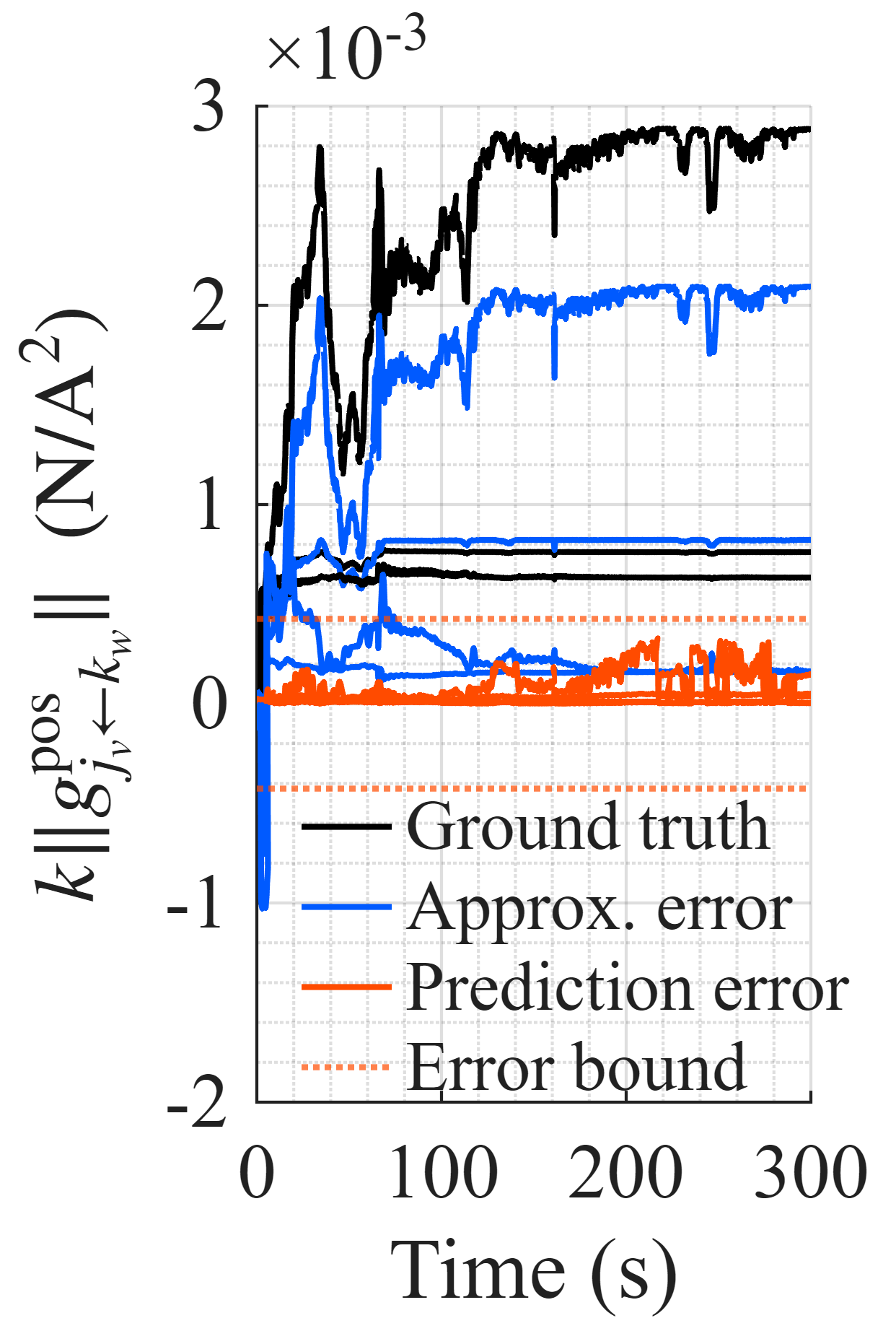}\label{CGL_fig:experimental_prediction_position_result}}
\end{minipage}
\begin{minipage}[b]{0.46\FigWidth}
\centering
  \subfloat[Prediction error of $\|g^{\mathrm{rot}}\|$\label{CGL_fig:learning_control}]{\includegraphics[width=1.05\linewidth]{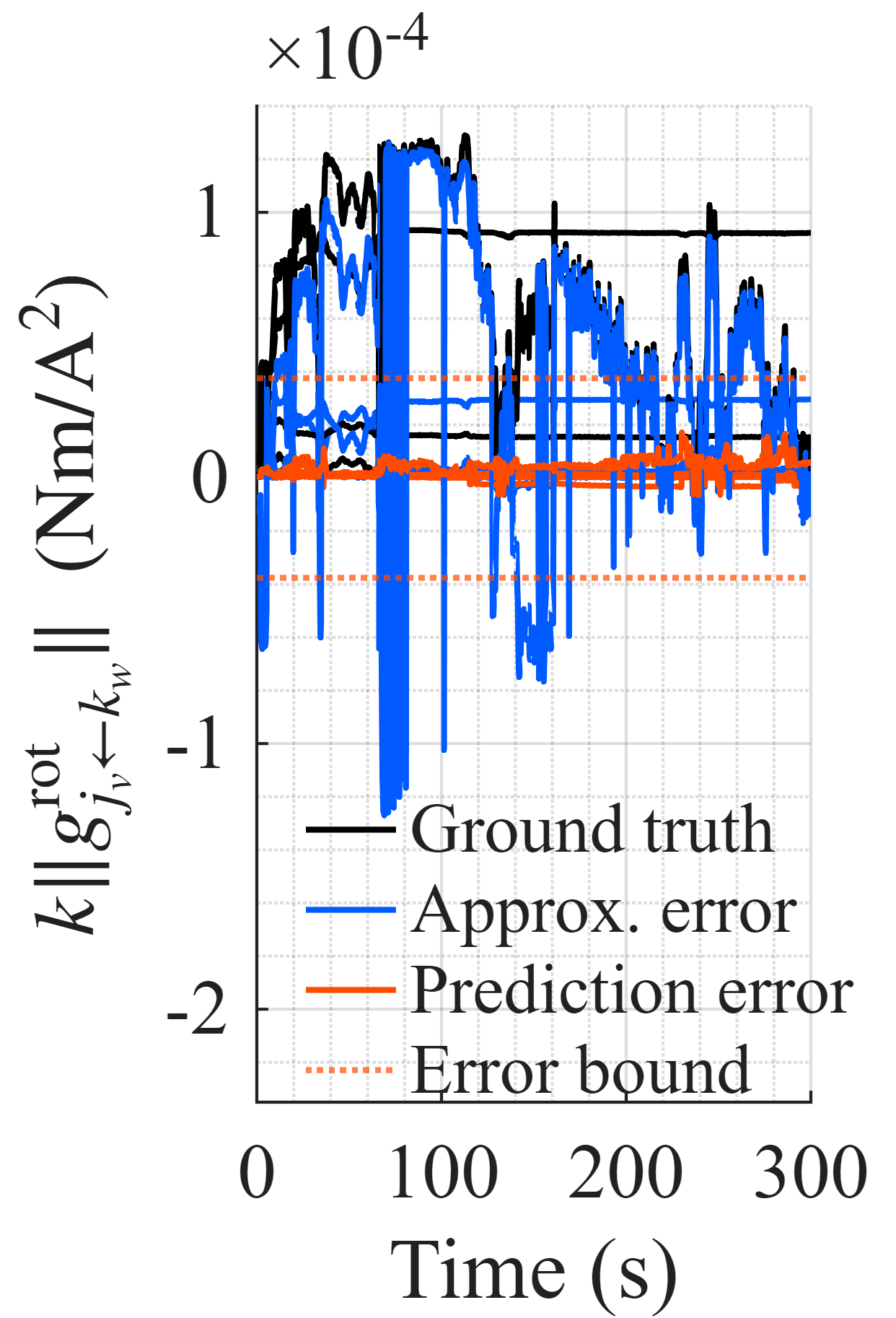}\label{CGL_fig:experimental_prediction_attitude_result}}  
\end{minipage}
\caption{Experimental results of two‐satellite docking control using the learned magnetic interaction model on a microgravity testbed \cite{takahashi2025noda_mmh}. With the constant $k\approx 2.205e^{-7}$ [H/m] applied, the y-axis values in Figs.~\ref{CGL_fig:experimental_prediction_position_result} and \ref{CGL_fig:experimental_prediction_attitude_result} indicate the coefficients relating the product of the two coil currents to the resulting force and torque.}
\label{CGL_experiment_formation_and_angle_control_results}
\end{figure}

We validate our trained magnetic interaction model $\mathfrak{g}$ through a ground testbed for a docking situation. We briefly summarize the experimental setup; please see the detailed description in \cite{takahashi2025noda_mmh}. The testbed MTQ has two-axis coils to generate a magnetic field in the horizontal plane, and the second coil uses an iron core. As shown in Fig.~\ref{CGL_fig:setup}, two MTQ mounts are on a linear air track and a single-axis air bearing, respectively. We measure relative position using a ToF sensor and absolute attitude using an AR marker-based camera pose estimation. The Pulse Per Second (PPS) signal from GPS enables time synchronization for AC magnetic control with an accuracy of 0.1 ms. Along with the assumption~\ref{CGL_target_chaser_assumption}, we assume the MTQ on a linear track has RWs to fix its attitude, and only the attitude of the MTQ on air-bearing is controlled. 

To compensate for the microgravity caused by linear track distortions, we design a PID controller for position and attitude control, updating its command values every 187.5 ms \textcolor{black}{with PID gains: position control $(K_p,K_i,K_d)=(5e^{-2},\,5e^{-3},\,5.2e^{-1})$, and attitude control $(K_p,K_i,K_d)=(1.3e^{-3},\,2e^{-5},\,8e^{-3})$.} Figures~\ref{CGL_fig:initial} and \ref{CGL_fig:final} show the initial and target conditions on the linear air track and single-axis air bearing in Fig.~\ref{CGL_fig:setup}. We calculate control current to realize the controller command value using the far-field model ``dipole'' and the learned model ``cgl'' trained in subsection~\ref{CGL_Offline_Training}. We also conduct the experiment on the learned model $\mathfrak{g}$ with spectral normalization ``cgl SN'' and without it ``cgl''. For every dipole allocation phase, the current amplitudes are calculated by the decentralized current allocation on (\ref{CGL_decentralized_current_allocation}). Unlike the assumption~\ref{CGL_fixed_current}, we normalize the norm of sine and cosine amplitudes of two coils to satisfy the current limitation of the testbed. 

Figures~\ref{CGL_fig:experimental_position_result} and \ref{CGL_fig:experimental_attitude_result} show the experimental results of relative position and absolute attitude. Although their control commands are the same, far-field model-based control cannot stabilize the states at the target in some cases; otherwise, it increases steady-position errors. Its reason is illustrated in Figs.~\ref{CGL_fig:experimental_prediction_position_result} and \ref{CGL_fig:experimental_prediction_attitude_result} that show the estimation results of coil geometry information vectors through each model with $k=\gamma_{\mu/c}^2\ {\mu_0}/{(8\pi A^2)}\approx 2.205e^{-7}$ and $\gamma_{\mu/c}\approx 2.1$ \cite{takahashi2025noda_mmh} [m$^2$] is the ratio between the dipole and current. The black line indicates the ground truth of coil geometry information vectors between each axis, and the blue line indicates the error between truth and the dipole approximation model prediction. As the distance decreases, the approximation becomes invalid and can result in large errors accompanied by sign reversals, which is consistent with the results presented in subsection~\ref{CGL_numerical_validation}. In contrast, the learning-based model ``cgl SN'' accurately predicts the ground-truth data, and the red curve represents the error between truth and the learned model prediction. The theoretical upper bound of ``cgl SN'' derived in Theorem~\ref{CGL_theorem_Certified_Error_Bounds} estimates the prediction error bound correctly. As a result, the learning-based magnetic field control successfully achieves rapid state convergence for proximity operation. Comparing ``cgl SN'' and ``cgl'', ``cgl SN'' achieves comparable—or even more stable—performance even though ``cgl SN'' exhibits relatively larger training errors as described in subsection~\ref{CGL_Offline_Training}. 
\section{Conclusions}
This paper proposes a learning-based magnetic interaction modeling approach, along with its validation, that mimics the exact model. Our approximation model reduces the computational burden of the exact magnetic model while minimizing the far-field approximation error in proximity. We first extend a previous decentralized current calculation, which does not require inter-agent communication, to account for the precise model and coil offset. Then, we demonstrate that the traditional approximation loses accuracy in proximity, leading to unstable motion, such as collisions. In contrast, our learning-based magnetic field model simultaneously controls relative distance and attitude with a low computational load and high stability. Its applications include several sectors that require exceptional accuracy and responsiveness. Especially, fuel-free docking is a key technology for in-space assembly, station resupply, sample-return, and large-scale swarm formation. 
\section*{Appendix}
\appendices
\subsection{Proof: Theorem~\ref{CGL_theorem_Certified_Error_Bounds} (Certified Error Bounds)}
\label{CGL_proof_Certified_Error_Bounds}
\begin{proof}
We first derive blockwise Lipschitz constants $L_{g_{\mathrm{pos},\mathrm{rot}}}^{{\mathrm{pos},\mathrm{rot}}}$ of the true function. We define the vectors $g_{\mathrm{pos,rot}}\in\mathbb{R}^3$ associated with the kernel of the double integral in (\ref{CGL_circulant_integration_term}) and an arbitrary vector $f_{(r)}=r+C^{k/j}R_j-R_k$ with $\|f_{(r)}\|\geq r_{\min}$:
$$
\left\{
\begin{aligned}
g_{\mathrm{pos}}&=K_{(r,\mathrm{d}l_k)}\times \mathrm{d}l_j\\
g_{\mathrm{rot}}&=[R_j\times K_{(r,\mathrm{d}l_k)}]\times \mathrm{d}l_j
\end{aligned}
\right.,\ 
K_{(r,\mathrm{d}l_k)}=\frac{f_{(r)}\times \mathrm{d}l_k}{\|f_{(r)}\|^3}\in\mathbb{R}^3
$$
where $R_j=a_{\mathrm{NN}}[\cos\varphi_j,\sin\varphi_j,0]^\top$ and $r\in\mathbb{R}^3$. Here, the Jacobian of $K$ satisfies 
$\|\partial K/\partial r\|_2\leq {\|I-3\hat{f}_{(r)}\hat{f}_{(r)}^\top\|\|[\mathrm{d}l_k]_\times\|}/{\|f_{(r)}\|^3}
\leq {2\|\mathrm{d}l_k\|}/{r_{\min}^3}$ where we use $\hat r=r/\|r\|$, $\|[\mathrm{d}l_k]_\times\|=\|\mathrm{d}l_k\|$, and 
$\lambda(I-3\hat{f}_r\hat{f}_r^\top)=\{1,1,-2\}$. Applying $\oint \|\mathrm{d}l_j\|=\oint \|\mathrm{d}l_k\|= 2\pi a_{\mathrm{NN}}$ yields the block-wise Lipschitz constants of $\bm g^{\mathrm{pos}}$ as 
$$
L_{\bm g_{\mathrm{pos}}}^{\mathrm{pos}}\leq L_0,\quad
L_{\bm g_{\mathrm{rot}}}^{\mathrm{pos}}\leq a_{\mathrm{NN}}L_0,\quad L_0={2(2\pi a_{\mathrm{NN}})^2}/{r_{\min}^3}
$$
where $\|K\|\le {\|\mathrm{d}l_k\|}/{\|f_{(r)}\|^2}\leq \|\mathrm{d}l_k\|/r_{\min}^2$ and $\|{\partial \mathrm{d}l_j}/{\partial \mathrm{n}_j}\|\leq \|\mathrm{d}l_j\|$. We next derive blockwise attitude Lipschitz constants. Let $\delta\mathrm{n}_j$ be denoted as the infinitesimal variation of the coil unit direction $\mathrm{n}_j$ in subsection~\ref{CGL_Reduced_Order_Sample_Collection_and_Learning}. Then, the first-order variations with  respect to $\mathrm{n}_j$ are  
$\|\delta g_{\mathrm{pos}}\|=\|[(\frac{\partial K}{\partial f_{(r)}}\frac{\partial f_{(r)}}{\partial \mathrm{n}_j}\delta \mathrm{n}_j)\times \mathrm{d}l_j+K\times \frac{\partial \mathrm{d}l_j}{\partial \mathrm{n}_j}] \delta \mathrm{n}_j\|\leq ( \frac{2a_{\mathrm{NN}}}{r_{\min}^3} + \frac{1}{r_{\min}^2})\|\mathrm{d}l_j\|\|\mathrm{d}l_k\|\|\delta \mathrm{n}_j\|$ and $\|\delta g_{\mathrm{rot}}\|=\|(\frac{\partial R_j}{\partial \mathrm{n}_j}\delta \mathrm{n}_j \times K+R_j\times \frac{\partial K}{\partial f_{(r)}} \frac{\partial f_{(r)}}{\partial \mathrm{n}_j}\delta \mathrm{n}_j)\times \mathrm{d}l_j+(R_j\times K)\times \frac{\partial \mathrm{d}l_j}{\partial \mathrm{n}_j}\delta \mathrm{n}_j\|
\leq(\frac{2a_{\mathrm{NN}}}{r_{\min}^2}+\frac{2a_{\mathrm{NN}}^2}{r_{\min}^3})\|\mathrm{d}l_j\|\|\mathrm{d}l_k\|\|\delta \mathrm{n}_j\|$
where $\|{\partial K}/{\partial f_{(r)}}\|
 \le 2\|\mathrm{d}l_k\|/r_{\min}^3$, $\|[R_{j_v}]_\times \| \leq a_{\mathrm{NN}}$, $
\|{\partial f_{(r)}}/{\partial \mathrm{n}_j}\|\leq a_{\mathrm{NN}}$, $\|{\partial R_j}/{\partial \mathrm{n}_j}\|\leq a_{\mathrm{NN}}$, and $\|{\partial K}/{\partial \mathrm{d}l_k}\|\leq {\|r\|^{-2}}$. Since the Jacobian $\partial \mathrm{n}/\partial \phi\in\mathbb{R}^{3\times 2}$
satisfies $\|(\partial \mathrm{n}/\partial \phi)^\top\partial \mathrm{n}/\partial \phi\|_2=\|\mathrm{diag}(\sin^2\phi_{(2)},1)\|_2=1$, we obtain $\|\delta\mathrm{ n}_j\|\leq\|\delta\phi_j\|$. Integrating $\|\mathrm{d}l_{j,k}\|$ over $\theta_{j,k}$ yields blockwise Lipschitz constants with respect to $\phi$ as $$
L_{\bm g_{\mathrm{pos}}}^{\mathrm{rot}}
\leq \left(a_{\mathrm{NN}}+r_{\min}/2\right)L_0,\quad 
L_{\bm g_{\mathrm{rot}}}^{\mathrm{rot}}
\leq \left(a_{\mathrm{NN}}+r_{\min}\right){{a_{\mathrm{NN}}}}L_0.
$$

We next derive the upper bound of $\rho_{\mathcal{S}_{\{(a_{\mathrm{NN}},r),(a_{\mathrm{NN}},\phi)\}}}$. Since $N$ points are optimally placed over the entire region, each point covers $\mathrm{Vol}(S_{\{(a_{\mathrm{NN}},r),(a_{\mathrm{NN}},\phi)\}})/\sqrt{N}$. Moreover, $n$-dimensional ball of radius $r$ generally has volume $r^n{\pi^{n/2}}/{\Gamma(1+n/2)}$ where $\Gamma(\cdot)$ denotes the gamma function. These observations yield the deterministic upper bound: 
$\rho_{\mathcal{S}_{\{(a_{\mathrm{NN}},r),(a_{\mathrm{NN}},\phi)\}}}\leq\frac{\mathrm{Vol}(\mathcal{S}_{\{(a_{\mathrm{NN}},r),(a_{\mathrm{NN}},\phi)\}})}{\sqrt{N}{\pi^{n/2}}/{\Gamma(\tfrac{n}{2}+1)} })^{1/n}$
where $\rho_{\mathcal{S}}\leq ({{\mathrm{Vol}(\mathcal{S})}/{(\pi \sqrt{N})}})^{1/2}$ for $n=2$
, $\mathrm{Vol}(\mathcal{S}_{(a_{\mathrm{NN}},r)})=({\pi}/{4})(\overline{r}^2-\underline{r}^2)$ and $\mathrm{Vol}(\mathcal{S}_{(a_{\mathrm{NN}},\phi)})=\pi(2\pi)$. These results conclude (\ref{CGL_eq_certified_error_bound}).

The prediction error for a coil radius $a_{\mathrm{inf}}$ is bounded as in (\ref{CGL_prediction_error_lipchitz_different_coil}) for 
$a_{\mathrm{inf}}=\gamma a_{\mathrm{NN}}$ in Theorem~\ref{CGL_Generalization_CGL}.
\end{proof}
\subsection{Earth Orbital and Attitude Motion Simulation}
\label{CGL_RW_attitude}
The relative motion in a circular orbit and the attitude dynamics of two satellites with RWs \cite{takahashi2022kinematics} are
\begin{equation*}
\begin{aligned}
&\ddot{x}+2 \omega_o \dot{y}-3 \omega_o^2 x  ={u_x},\quad\ddot{y}-2 \omega_o \dot{x} ={u_y}, \quad\ddot{z}+\omega_o^2 z ={u_z}\\
&J  \dot\omega+\omega \times\left(J  \omega+h_{R W}\right) =-\tau_{R W},\quad \dot{h}_{R W}=\tau_{R W}
\end{aligned}
\end{equation*}
where $u_{x,y,z}$ is acceleration per unit mass along each axis, $\omega_o\approx 1.1e^{-3}$ denotes the orbital frequency, $J$ is the moment of inertia, $\omega\in\mathbb{R}^{3}$ is the angular velocity, and $h_{R W}$ and $\tau_{R W}$ are the RWs angular momentum and torque, respectively.
\section*{Acknowledgment}
\noindent
The authors would like to thank Yoichi Tomioka at the University of Aizu for technical discussion, Atsuki Ochi at Interstellar Technologies Inc. for controller gain tuning, and the anonymous reviewers for their valuable comments.
\bibliographystyle{IEEEtran}
\bibliography{references}
\end{document}